\documentclass[preprint,12pt,number]{elsarticle}

\usepackage{graphicx}  
\usepackage{tikz}      
\usetikzlibrary{shapes.geometric, arrows} 

\usepackage{tikz}
\usetikzlibrary{positioning,shapes.geometric,arrows.meta,calc,decorations.pathmorphing}
\usepackage{lineno}
\usepackage{graphicx}  
\usepackage{pgfplots}  
\pgfplotsset{compat=1.18} 
\usepackage{tikz}
\usetikzlibrary{arrows.meta, positioning}
\usepackage{amsmath, amssymb, amsfonts}
\usepackage{graphicx}
\usepackage{hyperref}
\usepackage{listings}
\usepackage{xcolor}
\usepackage{geometry}
\usepackage{algorithm}
\usepackage{algpseudocode}
\usepackage{amsmath}
\usepackage{caption}
\usepackage{longtable}
\usepackage{booktabs}
 \usepackage[table,xcdraw]{xcolor}
\usepackage{tikz}
\usetikzlibrary{shapes.geometric, arrows}
\usepackage{amssymb}
\usepackage{amsmath}
\usepackage{hyperref}
\usepackage{natbib}
\tikzstyle{startstop} = [rectangle, rounded corners, minimum width=3cm, minimum height=1cm,text centered, draw=black, fill=red!30]
\tikzstyle{process} = [rectangle, minimum width=3cm, minimum height=1cm, text centered, draw=black, fill=orange!30]
\tikzstyle{block} = [rectangle, minimum width=3.5cm, minimum height=1cm, text centered, draw=black, fill=green!30]
\tikzstyle{arrow} = [thick,->,>=stealth]
\usepackage{tikz}
\usetikzlibrary{shapes.geometric,shadows, arrows}

\tikzstyle{startstop} = [rectangle, rounded corners, minimum width=3cm, minimum height=1cm,text centered, draw=black]
\tikzstyle{process} = [rectangle, minimum width=3cm, minimum height=1cm, text centered, draw=black]
\tikzstyle{decision} = [diamond, minimum width=3cm, minimum height=1cm, text centered, draw=black]
\tikzstyle{arrow} = [thick,->,>=stealth]
\usepackage{tikz}
\usetikzlibrary{shapes.geometric, arrows, positioning}

\tikzstyle{startstop} = [rectangle, rounded corners, minimum width=3cm, minimum height=1cm, text centered, draw=black, fill=blue!20]
\tikzstyle{process} = [rectangle, minimum width=3cm, minimum height=1cm, text centered, draw=black, fill=green!20]
\tikzstyle{decision} = [rectangle, minimum width=1cm, minimum height=1cm, text centered, draw=black, fill=yellow!20]
\tikzstyle{arrow} = [thick,->,>=stealth]

\journal{}

\begin{document}

\begin{frontmatter}



 \title{HybridSOMSpikeNet: A Deep Model with Differentiable Soft Self-Organizing Maps and Spiking Dynamics for Waste Classification}
\author[1]{Debojyoti Ghosh\footnote{Corresponding Author: debojyoyi07.dg@gmail.com}}

\author[1]{Adrijit Goswami}

\affiliation[1]{organization={Department of Mathematics, Indian Institute of Technology Kharagpur}, 
                addressline={Kharagpur}, 
                postcode={721302}, 
                state={West Bengal}, 
                country={India}}

\begin{abstract}
Accurate waste classification is vital for achieving sustainable waste management and
reducing the environmental footprint of urbanization. Misclassification of recyclable
materials contributes to landfill accumulation, inefficient recycling, and increased
greenhouse gas emissions. To address these issues, this study introduces
\textit{HybridSOMSpikeNet}, a hybrid deep learning framework that integrates convolutional
feature extraction, differentiable self-organization, and spiking-inspired temporal
processing to enable intelligent and energy-efficient waste classification.
The proposed model employs a pre-trained ResNet-152 backbone to extract deep spatial
representations, followed by a Differentiable Soft Self-Organizing Map (Soft-SOM) that
enhances topological clustering and interpretability. A spiking neural head accumulates
temporal activations over discrete time steps, improving robustness and generalization.
Trained on a ten-class waste dataset, \textit{HybridSOMSpikeNet} achieved a test accuracy of 97.39\%, outperforming several state-of-the-art architectures while maintaining a
lightweight computational profile suitable for real-world deployment.
Beyond its technical innovations, the framework provides tangible environmental
benefits. By enabling precise and automated waste segregation, it supports higher
recycling efficiency, reduces contamination in recyclable streams, and minimizes the
ecological and operational costs of waste processing. The approach aligns with global
sustainability priorities, particularly the United Nations Sustainable Development Goals
(SDG\,11 and SDG\,12), by contributing to cleaner cities, circular economy initiatives, and
intelligent environmental management systems.
\end{abstract}


\end{frontmatter}

\section{Introduction}

Rapid urbanization, industrial expansion, and population growth have led to an
unprecedented rise in global waste generation. According to the World Bank's
\textit{What a Waste 2.0} report, more than 2.2 billion tonnes of solid waste are produced
annually worldwide, with projections reaching 3.4 billion tonnes by 2050. The
inefficient sorting and recycling of this waste contribute significantly to
environmental degradation, including soil and water contamination, greenhouse gas
emissions, and excessive landfill accumulation. Effective waste classification and
segregation at the source are therefore crucial for advancing sustainable waste
management and achieving the targets of the circular economy.
Recent advances in artificial intelligence (AI) and computer vision have opened new
opportunities for addressing these challenges. Deep learning, in particular, has proven
effective in recognizing complex visual patterns, making it well suited for automated
waste identification. However, existing models often emphasize accuracy while
overlooking the environmental implications of computational cost, deployment
efficiency, and adaptability to evolving waste streams. To make AI a truly sustainable
tool for waste management, models must balance predictive performance with
interpretability, low energy demand, and integration potential within smart recycling
infrastructure.

Accurate and efficient waste categorization is crucial for enhancing recycling efficiency and promote environmental sustainability. Conventional human sorting techniques are often laborious and susceptible to errors, and can expose workers to health risks \cite{1}. Consequently, deep learning-based automated systems have arisen as an achievable choice to enhance waste classification processes.

Convolutional Neural Networks have shown strong capabilities in learning meaningful features from complex visual data. However, garbage classification continues to be an impressive effort because of inter-class similarities, intra-class variations, and cluttered backgrounds. Achieving high classification accuracy while maintaining computational efficiency continues to be a key concern, especially in real-world applications where both performance and speed are critical.


Among various deep learning methods, CNNs have proven highly effective in recognition of images roles, such as automatic waste classification \cite{2}. These networks can learn to identify and categorize waste types such as organic matter, recyclables, and hazardous materials without human supervision. Their strength lies in their capacity to analyze substantial volumes of images and extract hierarchical features that capture both high-level and low-level image characteristics \cite{3}.

Despite these advantages, numerous problems exist for developing robust deep learning algorithms for garbage categorization.  A significant difficulty is the limited availability of large, diverse, and accurately labeled datasets \cite{4}. Waste images can differ greatly in texture, shape, and color, which makes it is challenging for a single model to generalize across all categories \cite{5}. In addition, data labeling for waste classification is often limited, restricting dataset comprehensiveness. To address these limitations, techniques such as data augmentation, transfer learning, and synthetic data generation are commonly used to improve model robustness and overall performance \cite{6}.

Another challenge arises from the diversity of waste materials, each with unique physical and visual characteristics. Categories such as plastics, metals, paper, and organic waste differ in appearance and composition, making classification more complex \cite{7}. The presence of mixed or cluttered waste adds further difficulty. To tackle this, hybrid deep learning approaches that combine different architectures or integrate additional sensory data have been explored \cite{8}. For instance, combining CNNs with RNNs allows models to leverage both temporal and spatial information, leading to better classification accuracy \cite{9}.

Furthermore, the computational demands of deep learning models can make deployment difficult in environments with limited resources such as intelligent trash or embedded devices \cite{10}. Model optimization strategies including pruning, quantization, and lightweight architectures such as MobileNet have been introduced to reduce computational costs while maintaining accuracy \cite{11}. Additionally, combining CNNs with traditional machine learning classifiers like SVM has been investigated to balance efficiency and performance \cite{1}.

A combination of deep learning with Internet of Things (IoT) technologies has also created new opportunities in intelligent waste management. Smart bins integrated with sensing along with deep neural networks can autonomously identify and categorize garbage in real time \cite{13}. This combination enables adaptive, data-driven waste management systems that improve sorting, collection, and recycling processes, supporting sustainability and operational efficiency.

However, scaling deep learning-based garbage classification systems for real-world use remains challenging. Models must adapt to new or evolving waste categories and variations in data distribution over time. Transfer learning has been widely explored as a solution, allowing pre-trained models to be fine-tuned on smaller, updated datasets while reducing the need for retraining from scratch \cite{14}. Moreover, model transparency and interpretability have become increasingly important, especially in public waste management systems. Explainable AI (XAI) techniques provide insights into how models make decisions, building trust among users and stakeholders \cite{15}. This interpretability also supports model refinement and better adaptation to real-world conditions.

In this paper, we propose \textbf{HybridSOMSpikeNet}, a novel hybrid deep learning framework that integrates static feature learning, unsupervised topological clustering, and temporal spike-based processing. The proposed architecture employs a pre-trained ResNet152 model as a feature extractor to capture rich spatial features. These features are then passed to a Soft Self-Organizing Map (Soft-SOM) layer, which introduces an unsupervised learning mechanism to enhance class separability and capture topological relationships among data samples. Finally, a spiking neural network (SNN)-based classification head models temporal spike dynamics to produce the final predictions.

By separating visual representation, clustering, and temporal reasoning components, HybridSOMSpikeNet offers an energy-efficient, and robust approach to waste image classification. Extensive experiments conducted on a custom multi-class waste dataset demonstrate that the proposed model surpasses several state-of-the-art CNN architectures in accuracy and generalization, while maintaining a lower computational footprint suitable for deployment on low-power devices. This work contributes to the growing research at the intersection of symbolic learning, deep visual representation, and neuromorphic computing, providing a promising direction for sustainable and intelligent environmental AI systems.

The subsequent sections of this work are organized as followed.  Section~\ref{LS} provides an exhaustive analysis of relevant research and foundational research on garbage classification, self-organizing maps (SOM), and spiking neural networks (SNNs). Section~\ref{DP} outlines the dataset characteristics and the preprocessing techniques applied to ensure robust training. In Section~\ref{PA}, we detail the proposed HybridSOMSpikeNet architecture, emphasizing the integration of SOM-based unsupervised learning with neuromorphic SNN modules and deep feature extractors. Section~\ref{TS} describes the training strategy adopted, including learning rate scheduling, synaptic updates, and SOM-SNN co-adaptation. Section~\ref{Exp} elaborates on the experimental setup, covering evaluation metrics, training configurations, and baseline comparisons. Section~\ref{RD} provides an in-depth analysis of the experimental results, including accuracy, efficiency, and interpretability metrics. Finally, Section~\ref{CFW} summarizes the work and presents prospective avenues for further research, including energy-efficient training, model compression, and real-world deployment in intelligent waste disposal systems.

\section{Literature Survey}\label{LS}

The development of intelligent systems for waste classification has attracted significant attention due to the increasing demand for sustainable waste management solutions. The proposed HybridSOMSpikeNet model is inspired by advances in deep learning, self-organizing maps, and neuromorphic computing. This section categorizes the literature into four major domains: deep CNN-based classification, transfer learning and hybrid models, lightweight and optimized architectures, and neuro-inspired models, providing a comprehensive background for the proposed approach.

\subsection{Deep CNN-Based Models for Garbage Classification}
Deep Convolutional Neural Networks (CNNs) have demonstrated remarkable performance in image classification tasks. Pioneering works such as VGGNet \cite{simonyan2014very}, ResNet \cite{he2016deep}, EfficientNet \cite{tan2019efficientnet}, and NASNet \cite{zoph2018learning} have set benchmarks in computer vision and have been widely adopted in waste classification tasks. \cite{9435085} developed GarbageNet, a unified deep learning framework employing transfer learning and incremental learning to enhance recyclability-aware garbage classification.

Several studies applied these architectures to classify garbage images. \cite{1} developed a deep CNN-based system to automate waste segregation, while \cite{5} applied transfer learning using pre-trained CNNs like VGGNet for improved classification. ShuffleNet variants were optimized for mobile deployment in \cite{6}, and the MRS-YOLO model \cite{7} was introduced for real-time classification. \cite{madhavi2025swinconvnext} proposed SwinConvNeXt, a transformer-CNN hybrid for accurate continuous waste classification.

\subsection{Transfer Learning Techniques and Hybrid Models}
To overcome data scarcity and domain shift challenges, hybrid architectures and transfer learning have been employed. \cite{8} and \cite{9} explored transfer learning with CNNs for enhanced waste sorting accuracy. \cite{10} combined CNNs with deep reinforcement learning to enable adaptive classification. \cite{11} improved classification on occluded objects using attention-based convolution modules, while \cite{14} integrated transfer learning with lightweight CNNs for scalable, mobile-friendly systems.

Multi-task and multi-modal models were also explored. For example, \cite{1} proposed a hybrid CNN-autoencoder model for complex scenarios, and \cite{22} incorporated sensor data or metadata to improve robustness. Attention mechanisms \cite{25}, adversarial learning \cite{23}, and decision tree integrations \cite{24} further enhanced classification performance. \cite{10632565} proposed a multimodal dual cross-attention fusion strategy integrating image and audio modalities for robust autonomous garbage classification.

\subsection{Optimization and Lightweight Architectures}
Given the need for real-time and edge applications, optimization of deep models is crucial. \cite{16} proposed a lightweight CNN using depth-wise separable convolutions. \cite{19} optimized models for edge computing, while \cite{18} applied reinforcement learning to adapt CNNs to dynamic environments. \cite{21} reduced false positives and improved efficiency for deployment in smart city infrastructure.

The MobileNet family \cite{howard2017mobilenets} offers highly efficient architectures for resource-constrained settings, making them attractive for real-time garbage classification on edge devices.

\subsection{Neuro-Inspired and Unsupervised Learning Models}
To mimic human-like perception and adaptability, neuro-inspired and unsupervised models are gaining attention. The Self-Organizing Map (SOM), proposed by Kohonen \cite{kohonen1990self}, is a biologically inspired unsupervised learning algorithm known for its topological mapping capabilities, and has shown promise in clustering and feature extraction for high-dimensional data. \cite{10056283} introduced a prototype enhancement-based incremental evolution learning method using contrastive features to improve adaptability in urban garbage classification tasks.

Spiking Neural Networks (SNNs) are the third generation of neural networks, related mimicking biological neuron behavior through spike-based computation. Neuromorphic models \cite{roy2019towards}, and deep learning in SNNs \cite{tavanaei2019deep} offer energy-efficient, event-driven processing. Recent works \cite{sengupta2019going, diehl2015fast} extend deep architectures like VGG and ResNet into spike-based models using conversion or training techniques. These models demonstrate potential in real-time classification with low power consumption.

Integrating SOMs with SNNs, as proposed in HybridSOMSpikeNet, is motivated by the unsupervised feature learning capacity of SOMs and the biologically plausible, low-latency processing of SNNs. This fusion leverages the strengths of both topological self-organization and neuromorphic computation, making it ideal for scalable and efficient waste classification systems.

\subsection{Recent Innovations in Sustainable Waste Management}

Recent research has advanced waste classification and intelligent systems from multiple perspectives.  \cite{n1} enhanced fuzzy classification performance in high-dimensional feature spaces through feature combination optimization, while  \cite{n2} introduced a novel hazard classification model that integrates grey models with deep learning. In the context of sustainability,  \cite{n4} developed a predictive analysis framework for waste management in smart urban areas using edge computing and blockchain IoT. More recently, \cite{ghosh2025enhanced} presented an improved deep model for optimized garbage classification, specifically targeting smart waste management systems. Collectively, these studies highlight the growing synergy between deep learning, intelligent sensing, and sustainable waste management, motivating the development of more robust hybrid models such as our proposed HybridSOMSpikeNet. 
~\cite{WANG2025107350} proposed a convolutional transformer network for spatial-spectral fusion using contextual multi-head self-attention that combines convolutional and transformer modules to enhance both local and global feature extraction.
~\cite{XU2025107311} introduced a Dual Selective Fusion Transformer Network (DSFormer) that adaptively fuses spatial and spectral features across multiple receptive fields, achieving strong performance across several benchmark hyperspectral datasets.
Similarly,~\cite{Fu2025CTA} developed a network based on CNN-Transformer and Channel-Spatial Attention that effectively handles few-sample hyperspectral image classification by integrating attention mechanisms and hybrid feature extraction. \citet{ghosh2025enhanced} proposed an enhanced deep learning framework for efficient garbage classification in smart waste management systems, improving both accuracy and computational efficiency. \citet{n1} demonstrated that feature combination optimization can significantly boost the performance of high-dimensional fuzzy classification models. 
\citet{n2} developed a hybrid hazard classification model combining grey modeling with deep learning to improve predictive reliability.


 \subsection{Novelty and Effectiveness of HybridSOMSpikeNet}
\label{sec:novelty-effectiveness}

This work introduces, for the first time, a Differentiable Sof Self-Organizing Map, which extends the classical SOM by making the clustering operation fully differentiable and compatible with gradient-based learning. Unlike traditional SOMs that require separate, nongradient-based training, the Diff-SOM can be trained end-to-end alongside a deep convolutional backbone (ResNet-152) and a Spiking Neural Network (SNN) head. By enabling backpropagation through the SOM layer, our approach allows high-dimensional CNN features to be softly clustered into meaningful prototypes while simultaneously optimizing the downstream spiking classification. This is, to our knowledge, the first research to introduce a differentiable SOM, highlighting its potential to bridge topological feature learning and temporal spiking dynamics in a unified, trainable framework.\\[6pt]
The key novelties of this approach are as follows:
\begin{itemize}
   \item\textbf{Differentiable Soft SOM Layer:} Introduction of a fully differentiable hybrid architecture combining SOMs with spiking neurons. Traditional SOMs are unsupervised and nondifferentiable, which limits their integration into end-to-end deep learning pipelines. In this work, we introduce a soft, differentiable variant of SOM that allows gradient-based learning alongside CNN backbones. This enables the network to learn feature prototypes that are both semantically meaningful and optimized for downstream classification.
     \item\textbf{Integration with Spiking Head:} While spiking neural networks have primarily been explored in neuromorphic computing contexts, their integration with soft SOM representations for standard image classification tasks is novel. The spiking head aggregates soft SOM activations over multiple time steps, introducing temporal dynamics that improve robustness and generalization.
     \item\textbf{Hybrid CNN-SOM-SNN Architecture:} Existing literature typically explores CNNs, SOMs, or SNNs in isolation or in pairwise combinations. Our approach is the first to combine all three, resulting in a powerful hybrid model that leverages the feature extraction capability of deep CNNs, the topological clustering of SOMs, and the temporal dynamics of spiking neurons.
  \item\textbf{End-to-End Trainable:} Unlike conventional SOMs that require separate training, our differential soft SOM is fully integrated into the end-to-end learning pipeline. This allows the backbone, SOM, and spiking head to co-adapt during training, improving overall performance on complex classification tasks such as garbage recognition across diverse categories.
\end{itemize}

Overall, this work demonstrates that combining differentiable SOMs with spiking networks in a hybrid architecture not only yields strong performance but also opens a new direction for integrating topological learning and temporal dynamics in deep neural networks.

 \section{Dataset Overview and Preprocessing }\label{DP}
\label{sec:data_preprocessing}

The preprocessing pipeline for the waste classification dataset consists of two major steps: stratified dataset splitting, data transformation. Each step is designed to improve model robustness and generalization.



\subsection{Dataset Overview}

The data utilized in the present study was obtained from \href{https://www.kaggle.com/datasets/mostafaabla/garbage-classification/data}{Kaggle}\footnote{\url{https://www.kaggle.com/datasets/mostafaabla/garbage-classification/data}}, a widely recognized repository for datasets. The original dataset contained 12 distinct categories, including three classes representing different types of glass. For the purposes of this research, only the \textit{white-glass} class was retained, while the other glass-related categories were excluded.

A curated garbage classification dataset, denoted as $\mathcal{D}$, was constructed for this study, consisting of RGB images organized into $C = 10$ distinct waste categories:

\begin{equation*}
\mathcal{C} = \{\text{battery}, \text{biological}, \text{cardboard}, \text{clothes}, \text{metal}, \text{paper}, \text{plastic}, \text{shoes}, \text{trash}, \text{white-glass}\}.
\end{equation*}

\begin{figure}[h!]
\centering
\includegraphics[width=0.8\linewidth]{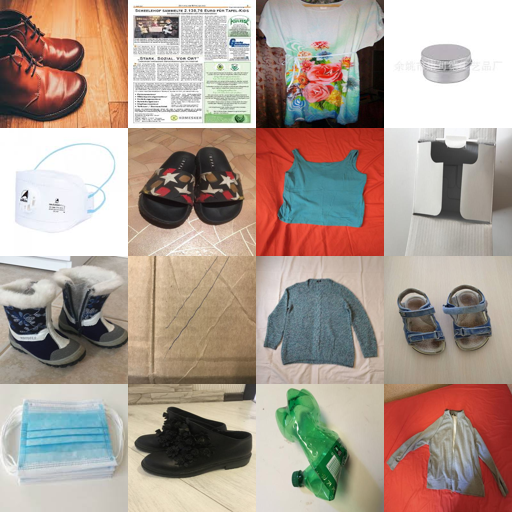}
\caption{Sample images from the curated garbage classification dataset.}
\label{fig:dataset_overview_16}
\end{figure}

The complete dataset comprises \textbf{14,279 images} distributed across these ten classes. Each category contains images of waste items captured under diverse lighting conditions, orientations, and backgrounds, providing a realistic and varied dataset suitable for training and evaluating deep learning-based waste classification models. The distribution of images per class is summarized in Table~\ref{tab:dataset_total_distribution}. A few sample images from the curated garbage classification dataset are shown in Figure~\ref{fig:dataset_overview_16}.

\begin{table}[h!]
\centering
\caption{Total number of images per class in the curated dataset.}
\begin{tabular}{l c}
\hline
\textbf{Category} & \textbf{Number of Images} \\
\hline
Battery        & 945  \\
Biological     & 985  \\
Cardboard      & 891  \\
Clothes        & 5,325 \\
Metal          & 769  \\
Paper          & 1,050 \\
Plastic        & 865  \\
Shoes          & 1,977 \\
Trash          & 697  \\
White-glass    & 775  \\
\hline
\textbf{Total} & \textbf{14,279} \\
\hline
\end{tabular}
\label{tab:dataset_total_distribution}
\end{table}

\subsection{Dataset Splitting Strategy}

The dataset $\mathcal{D} = \{(x_i, y_i)\}_{i=1}^N$, with $x_i$ an image and $y_i \in \{1,\ldots,10\}$, is split into training, validation, and test sets using \texttt{StratifiedShuffleSplit} to preserve class distributions.  

Formally,  
\[
\mathcal{D} = \mathcal{D}_{\text{train}} \cup \mathcal{D}_{\text{val}} \cup \mathcal{D}_{\text{test}}, \quad
|\mathcal{D}_{\text{train}}| \approx 70\%, \; |\mathcal{D}_{\text{val}}| \approx 15\%, \; |\mathcal{D}_{\text{test}}| \approx 15\%.
\]

Splitting is done in two steps: (1) train vs. temporary set, (2) temporary into validation and test. This ensures all subsets retain the original class proportions for fair training and evaluation.

\subsection{Data Transformations}
\label{sec:data_transformations}

Two separate transformation pipelines were applied for training and evaluation to improve model robustness and ensure consistent preprocessing.

     \subsubsection{Training Transformations:}  
    The training images are subjected to a series of stochastic augmentations to enhance resilience to variations in scale, orientation, illumination, and perspective. Specifically, each image undergoes:
    \begin{itemize}
        \item Randomly resized cropping to $224 \times 224$ pixels.
        \item Random horizontal flipping.
        \item Color jittering with brightness, contrast, saturation, and hue adjustments.
        \item Random rotations of up to $15^\circ$ combined with minor affine translations (up to 5\% in each direction).
        \item Random perspective distortion with a distortion scale of 0.3.
        \item Conversion to a tensor suitable for PyTorch processing.
        \item Normalization using ImageNet mean $\mu=[0.485,0.456,0.406]$ and standard deviation $\sigma=[0.229,0.224,0.225]$.
    \end{itemize}

    \subsubsection{Validation and Test Transformations:}  
    For evaluation, deterministic preprocessing ensures consistent input dimensions and color scaling. Each image is:
    \begin{itemize}
        \item Resized to $256 \times 256$ pixels.
        \item Center-cropped to $224 \times 224$ pixels.
        \item Converted to a tensor.
        \item Normalized using the same ImageNet mean and standard deviation as the training set.
    \end{itemize}

\subsection{Data Loader Configuration}

The dataset is efficiently loaded using PyTorch’s \texttt{ImageFolder} utility, which automatically assigns labels based on directory structure. 
After loading, the data is split into training, validation, and test subsets. 
Each subset is then converted into mini-batches through the \texttt{DataLoader} interface, as described in Equation~\ref{eq:dataloader_config}.

\begin{align}
\label{eq:dataloader_config}
    \mathcal{B}_{\text{train}} &= \texttt{DataLoader}(\mathcal{D}_{\text{train}}, \text{batch\_size}=32, \text{shuffle=True}), \notag \\
    \mathcal{B}_{\text{val}}   &= \texttt{DataLoader}(\mathcal{D}_{\text{val}}, \text{batch\_size}=32, \text{shuffle=False}), \notag \\
    \mathcal{B}_{\text{test}}  &= \texttt{DataLoader}(\mathcal{D}_{\text{test}}, \text{batch\_size}=32, \text{shuffle=False}).
\end{align}

As shown in Equation~\ref{eq:dataloader_config}, 
the training loader shuffles data at every epoch to improve generalization, 
while validation and test loaders preserve a fixed order for consistent evaluation.

This preprocessing pipeline ensures a high degree of data variability during training while preserving label consistency and evaluation integrity during validation and testing.

 \section{Proposed Architecture}\label{PA}

\subsection{Overview of HybridSOMSpikeNet}
\begin{figure}[h!]
\centering
\begin{tikzpicture}[node distance=1cm, every node/.style={font=\small}]
    \tikzstyle{block} = [rectangle, minimum width=7cm, minimum height=1.2cm, text centered, draw=black, fill=blue!20]
    \tikzstyle{arrow} = [thick,->,>=stealth]

    \node (input) [block, fill=orange!25] {Input Image};
    \node (resnet) [block, below=of input] {ResNet152 Feature Extraction};
    \node (som) [block, below=of resnet] {Soft SOM (128-D Clustering)};
    \node (snn) [block, below=of som] {Spiking Head (Temporal Aggregation)};
    \node (output) [block, fill=green!25, below=of snn] {Predicted Class};

    \draw [arrow] (input) -- (resnet);
    \draw [arrow] (resnet) -- (som);
    \draw [arrow] (som) -- (snn);
    \draw [arrow] (snn) -- (output);
\end{tikzpicture}
\caption{Diagram of HybridSOMSpikeNet architecture showing sequential processing from input to output.}
\label{fig:hybridsomspikenet}
\end{figure}
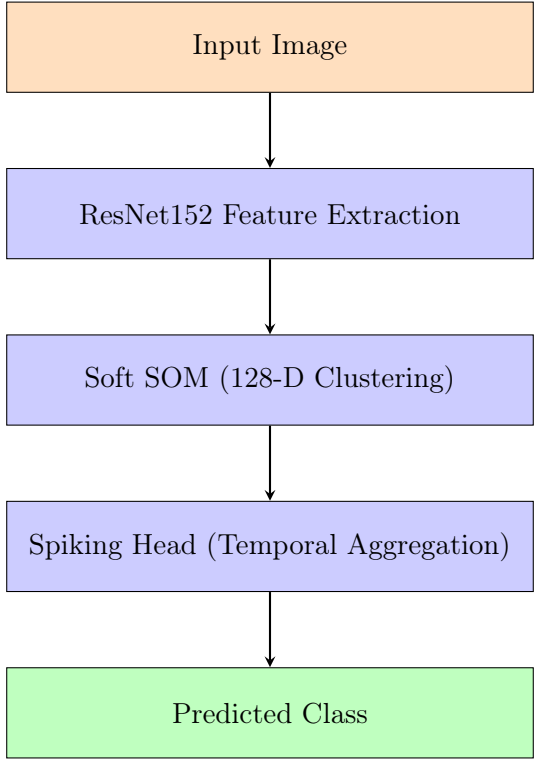

To aid in understanding the architecture, Figure~\ref{fig:hybridsomspikenet} provides a schematic diagram of the proposed HybridSOMSpikeNet model. It depicts the sequential flow from image input through the ResNet-based feature extractor, the soft clustering module (SSOL), and the temporally integrated SNN head, leading to the final classification. This layered design supports hierarchical feature learning, topological clustering, and biologically plausible inference in a lightweight, modular framework.

\subsection{Feature Extraction via Deep Residual Learning}
\label{sec:feature_extraction}

The first critical component of the proposed HybridSOMSpikeNet architecture is a robust feature extraction stage, which transforms raw input images into high-dimensional semantic embeddings. For this purpose, we utilize the well-established \textit{ResNet152} as a backbone convolutional neural network.

ResNet152, introduced in the seminal work by ~\cite{he2016deep}, is a deep residual network comprising 152 layers. Its key innovation is the use of identity-based \emph{shortcut connections}, which enable the propagation of gradients through deep architectures without suffering from vanishing or exploding effects. This architectural design allows for training very deep networks while maintaining strong generalization and convergence properties. The network consists of a series of residual blocks, each containing batch normalization, ReLU activations, and convolutional layers, connected through skip pathways that learn residual functions instead of direct mappings.

Let the input image be denoted as $\mathbf{X} \in \mathbb{R}^{H \times W \times 3}$. The image passes through multiple convolutional and residual stages of ResNet152. After the final convolutional stage, a \textit{global average pooling} (GAP) layer compresses spatial dimensions while preserving semantic richness. The output of this stage is a fixed-length embedding vector $\mathbf{F}$, as shown in \ref{F_embedding}:

\begin{equation}
\mathbf{F} = \text{ResNet152}_{\text{avgpool}}(\mathbf{X}), \quad \mathbf{F} \in \mathbb{R}^{2048}
\label{F_embedding}
\end{equation}
 This compact feature vector serves as a semantically rich representation of the original image and forms the input to the next stage in our pipeline: Soft Self-Organizing Map (Soft-SOM) clustering. By leveraging a pre-trained ResNet152, we significantly reduce the burden of training from scratch, while benefiting from transfer learning rooted in large-scale image datasets such as ImageNet.

\subsubsection{ Deep Feature Extraction from Image}
The first stage of the HybridSOMSpikeNet architecture involves extracting meaningful visual features from raw input images using a deep convolutional backbone. As outlined in Algorithm~\ref{alg:resnet_feature_extraction}, we employ a pre-trained ResNet152 model due to its proven robustness and depth. The input image is first normalized using ImageNet statistics and then passed through a sequence of convolutional and residual blocks. Following the final convolutional stage, a global average pooling (GAP) layer aggregates spatial information into a fixed-length feature representation. This 2048-dimensional vector encodes high-level semantic characteristics of the input and serves as input to the subsequent clustering layer. The use of ResNet152 leverages transfer learning, enabling efficient training even with limited labeled data while ensuring rich and generalizable representations.
\begin{algorithm}[h!]
\caption{Feature Extraction using ResNet152}
\label{alg:resnet_feature_extraction}
\begin{algorithmic}[1]
\Require Input image $\mathbf{X} \in \mathbb{R}^{H \times W \times 3}$
\Ensure Feature vector $\mathbf{F} \in \mathbb{R}^{2048}$
\State Normalize the input image $\mathbf{X}$ (mean subtraction, scaling)
\State Pass $\mathbf{X}$ through convolutional layers of ResNet152
\State Extract output from the global average pooling layer
\State Flatten the output to obtain $\mathbf{F}$
\State \Return $\mathbf{F}$
\end{algorithmic}
\end{algorithm}

\subsection{Soft Self-Organizing Layer (SSOL)}
\label{sec:ssol}

The Soft Self-Organizing Layer (SSOL) is a differentiable clustering module inspired by classical Self-Organizing Maps (SOMs), designed to operate within deep neural networks for end-to-end training. Traditional SOMs use hard competitive learning rules that are non-differentiable, limiting their integration into gradient-based learning frameworks. In contrast, SSOL introduces a soft assignment mechanism using distance-based softmax, enabling it to learn cluster structures while supporting backpropagation.

This layer plays a crucial role in HybridSOMSpikeNet by enhancing the topological discriminability of high-dimensional features extracted from a CNN backbone. It facilitates prototype-based clustering where each sample is softly associated with multiple cluster centers based on its proximity, encouraging smooth interpolation and robust representation learning.

\subsubsection{Mathematical Formulation}

Let $\mathbf{X} = [\mathbf{x}_1, \ldots, \mathbf{x}_N]^\top \in \mathbb{R}^{N \times d}$ represent a batch of $N$ input feature vectors of dimension $d$, and let $\mathbf{P} = [\mathbf{p}_1, \ldots, \mathbf{p}_K]^\top \in \mathbb{R}^{K \times d}$ denote $K$ learnable prototype vectors. The pairwise Euclidean distance matrix $\mathbf{D} \in \mathbb{R}^{N \times K}$ is computed as, in \ref{eq:distance}:

\begin{equation}
    \mathbf{D}_{ij} = \left\| \mathbf{x}_i - \mathbf{p}_j \right\|_2, 
    \quad \forall i \in [1, N],\ j \in [1, K]
    \label{eq:distance}
\end{equation}

To convert these distances into soft assignments, we apply a softmax function over the negative distances (to give higher weights to closer prototypes) in \ref{eq:soft_assign}:

\begin{equation}
    \mathbf{S}_{ij} = 
    \frac{\exp(-\mathbf{D}_{ij})}
         {\sum_{k=1}^{K} \exp(-\mathbf{D}_{ik})}
    \label{eq:soft_assign}
\end{equation}

The resulting matrix $\mathbf{S} \in \mathbb{R}^{N \times K}$ encodes the degree to which each input $\mathbf{x}_i$ is associated with each prototype $\mathbf{p}_j$. To promote generalization and mitigate overfitting, dropout regularization is optionally applied to $\mathbf{S}$ during training.

\subsubsection{Advantages of SSOL}
\begin{itemize}
\item\textbf{Differentiable Clustering:} SSOL supports full end-to-end training through backpropagation, unlike classical SOMs.
 \item\textbf{Topology Preservation:} The smooth assignment captures underlying topological structure without enforcing rigid cluster boundaries.
 \item\textbf{Dynamic Adaptivity:} Prototypes evolve with training, adapting to the current feature distribution and learning semantics.
\item\textbf{Energy-Efficient Integration:} Serves as a lightweight symbolic bridge between CNN features and the spike-based SNN head.
\end{itemize}

\subsubsection{Gradient Flow and Backpropagation}
\label{sec:ssol-backward}

To maintain full differentiability throughout the network, the Soft Self-Organizing Layer (SSOL) is constructed so that gradients can efficiently flow through both the distance computation and the softmax transformation. This section presents an analytical overview of the backward pass, which is crucial for updating the input embeddings and prototype vectors during training.

Assume a loss function $\mathcal{L}$ (such as cross-entropy or contrastive loss) is computed over the soft assignment matrix $\mathbf{S}$. The goal is to obtain gradients with respect to the input features $\mathbf{X}$ and the prototype set $\mathbf{P}$.

We begin by defining the following intermediate terms in \ref{eq:delta_and_distance}:
\begin{equation}
\delta_{ij} = \frac{\partial \mathcal{L}}{\partial \mathbf{S}_{ij}}, 
\qquad 
\mathbf{D}_{ij} = \| \mathbf{x}_i - \mathbf{p}_j \|_2
\label{eq:delta_and_distance}
\end{equation}

The partial derivatives of the softmax function with respect to the distances are provided in \ref{eq:softmax_derivative}:
\begin{equation}
\frac{\partial \mathbf{S}_{ij}}{\partial \mathbf{D}_{ij}} = -\mathbf{S}_{ij}(1 - \mathbf{S}_{ij}), 
\qquad 
\frac{\partial \mathbf{S}_{ij}}{\partial \mathbf{D}_{ik}} = \mathbf{S}_{ij}\mathbf{S}_{ik} 
\quad \text{for } j \neq k
\label{eq:softmax_derivative}
\end{equation}

Applying the chain rule, the gradient of the loss with respect to the input features $\mathbf{X}$ and the prototypes $\mathbf{P}$ can be computed as shown in \ref{eq:grad_x} and \ref{eq:grad_p}:
\begin{align}
\frac{\partial \mathcal{L}}{\partial \mathbf{x}_i} 
&= \sum_{j=1}^{K} \delta_{ij} 
\cdot 
\frac{\partial \mathbf{S}_{ij}}{\partial \mathbf{D}_{ij}} 
\cdot 
\frac{\partial \mathbf{D}_{ij}}{\partial \mathbf{x}_i}
= 
\sum_{j=1}^{K} \delta_{ij} 
\cdot 
\frac{\partial \mathbf{S}_{ij}}{\partial \mathbf{D}_{ij}} 
\cdot 
\frac{\mathbf{x}_i - \mathbf{p}_j}{\|\mathbf{x}_i - \mathbf{p}_j\|_2}
\label{eq:grad_x}
\\[2mm]
\frac{\partial \mathcal{L}}{\partial \mathbf{p}_j} 
&= 
-\sum_{i=1}^{N} \delta_{ij} 
\cdot 
\frac{\partial \mathbf{S}_{ij}}{\partial \mathbf{D}_{ij}} 
\cdot 
\frac{\mathbf{x}_i - \mathbf{p}_j}{\|\mathbf{x}_i - \mathbf{p}_j\|_2}
\label{eq:grad_p}
\end{align}

As illustrated in \ref{eq:delta_and_distance}--\ref{eq:grad_p}, these gradients enable the SSOL to be integrated seamlessly into modern deep learning pipelines. Standard optimizers such as Adam can then be applied to update both the input features and prototype vectors efficiently during training.

Algorithm~\ref{alg:ssol-forward} outlines the forward pass of the Soft Self-Organizing Layer. It is simple yet powerful, based entirely on distance computation and normalized exponential weighting.

\begin{algorithm}[h!]
\caption{Soft Self-Organizing Layer Forward Pass}
\label{alg:ssol-forward}
\begin{algorithmic}[1]
\Require Input batch $\mathbf{X} \in \mathbb{R}^{N \times d}$, Prototypes $\mathbf{P} \in \mathbb{R}^{K \times d}$, Dropout rate $p$
\Ensure Soft assignment matrix $\mathbf{S} \in \mathbb{R}^{N \times K}$

\State Compute Euclidean distance matrix: $\mathbf{D}_{ij} \gets \left\| \mathbf{x}_i - \mathbf{p}_j \right\|_2$
\State Apply softmax over negative distances: $\mathbf{S}_{ij} \gets \frac{\exp(-\mathbf{D}_{ij})}{\sum_{k=1}^K \exp(-\mathbf{D}_{ik})}$
\State Apply dropout: $\mathbf{S} \gets \texttt{Dropout}(\mathbf{S},\ p)$
\State \Return $\mathbf{S}$
\end{algorithmic}
\end{algorithm}

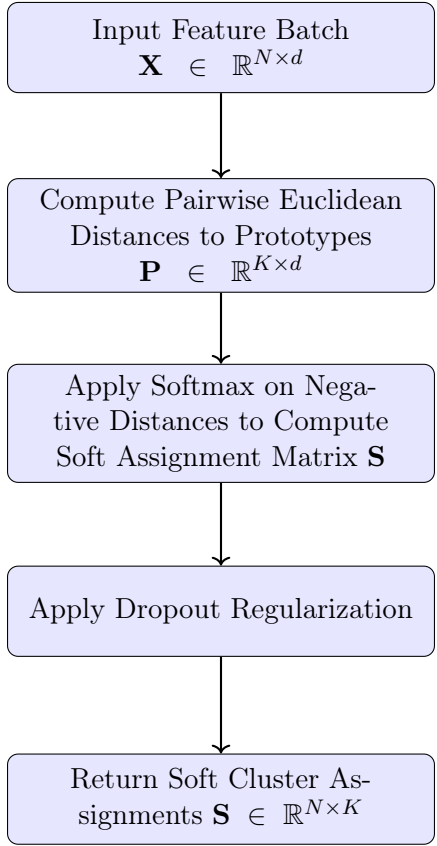
\begin{figure}[h!]
\centering
\begin{tikzpicture}[node distance=2.5cm, every node/.style={align=center, font=\small}]
\tikzstyle{block} = [rectangle, draw, fill=blue!10, text width=5.4cm, rounded corners, minimum height=1.2cm]
\tikzstyle{arrow} = [->, thick]

\node[block] (input) {Input Feature Batch \\ $\mathbf{X} \in \mathbb{R}^{N \times d}$};
\node[block, below of=input] (dist) {Compute Pairwise Euclidean Distances to Prototypes \\ $\mathbf{P} \in \mathbb{R}^{K \times d}$};
\node[block, below of=dist] (softmax) {Apply Softmax on Negative Distances to Compute \\ Soft Assignment Matrix $\mathbf{S}$};
\node[block, below of=softmax] (dropout) {Apply Dropout Regularization};
\node[block, below of=dropout] (output) {Return Soft Cluster Assignments $\mathbf{S} \in \mathbb{R}^{N \times K}$};

\draw[arrow] (input) -- (dist);
\draw[arrow] (dist) -- (softmax);
\draw[arrow] (softmax) -- (dropout);
\draw[arrow] (dropout) -- (output);
\end{tikzpicture}
\caption{Flowchart of the Soft Self-Organizing Layer (SSOL). The input batch is passed through a differentiable clustering process involving distance computation, soft assignment, and optional regularization.}
\label{fig:ssol-flowchart}
\end{figure}

Figure~\ref{fig:ssol-flowchart} visually summarizes the sequential operations of the Soft Self-Organizing Layer, from input reception to soft assignment generation.

\subsection{Spiking Head Module}
\label{sec:spiking_head}

The Spiking Head, a core component of the proposed HybridSOMSpikeNet architecture, serves as a biologically inspired temporal processing layer. It mimics the behavior of spiking neurons by integrating feature representations across multiple discrete time steps. Unlike traditional dense classifiers, this module exploits temporal accumulation to encode dynamic representations without the overhead of recurrent connections or explicit state maintenance.

\begin{figure*}[h!]
\centering
\begin{tikzpicture}[
    node distance=3.5cm,
    every node/.style={draw, align=center, font=\small, minimum width=1.2cm, minimum height=1cm},
    io/.style={fill=blue!10},
    layer/.style={fill=gray!10},
    logic/.style={fill=orange!15},
    result/.style={fill=green!15},
    arrow/.style={->, thick}
]
\node[io] (input) {Input Feature \\ $\mathbf{x}$};
\node[layer, right of=input] (fc1) {Linear Layer \\ $\mathbf{W}_1 \mathbf{x} + \mathbf{b}_1$};
\node[layer, right of=fc1] (bn1) {BatchNorm1};
\node[layer, right of=bn1] (relu1) {ReLU \\ $\mathbf{h}_t$};
\node[layer, right of=relu1] (fc2) {Linear Layer \\ $\mathbf{W}_2 \mathbf{h}_t + \mathbf{b}_2$};
\node[layer, below of=fc2] (bn2) {BatchNorm2};
\node[layer, left of=bn2] (relu2) {ReLU \\ $\mathbf{o}_t$};
\node[logic, left of=relu2] (accum) {Accumulate: \\ $\mathbf{M} += \mathbf{o}_t$};
\node[draw=none, below=of accum, font=\scriptsize] (loop) {Repeat for $t = 1 \dots T$};
\node[layer, left of=accum] (avg) {Average over\\Time:  $\mathbf{M} \gets \mathbf{M} / T$};
\node[result, below of=avg] (output) {Output Membrane \\Potential  $\mathbf{M}$};

\draw[arrow] (input) -- (fc1);
\draw[arrow] (fc1) -- (bn1);
\draw[arrow] (bn1) -- (relu1);
\draw[arrow] (relu1) -- (fc2);
\draw[arrow] (fc2) -- (bn2);
\draw[arrow] (bn2) -- (relu2);
\draw[arrow] (relu2) -- (accum);
\draw[->, dashed, dash pattern=on 2pt off 2pt] (accum.south) -- ++(0,0) -- ++(0,0) -- (loop);

\draw[arrow] (accum) -- (avg);
\draw[arrow] (avg) -- (output);
\end{tikzpicture}
\caption{Flowchart of the Spiking Head module.}
\label{fig:spiking_head_flow}
\end{figure*}
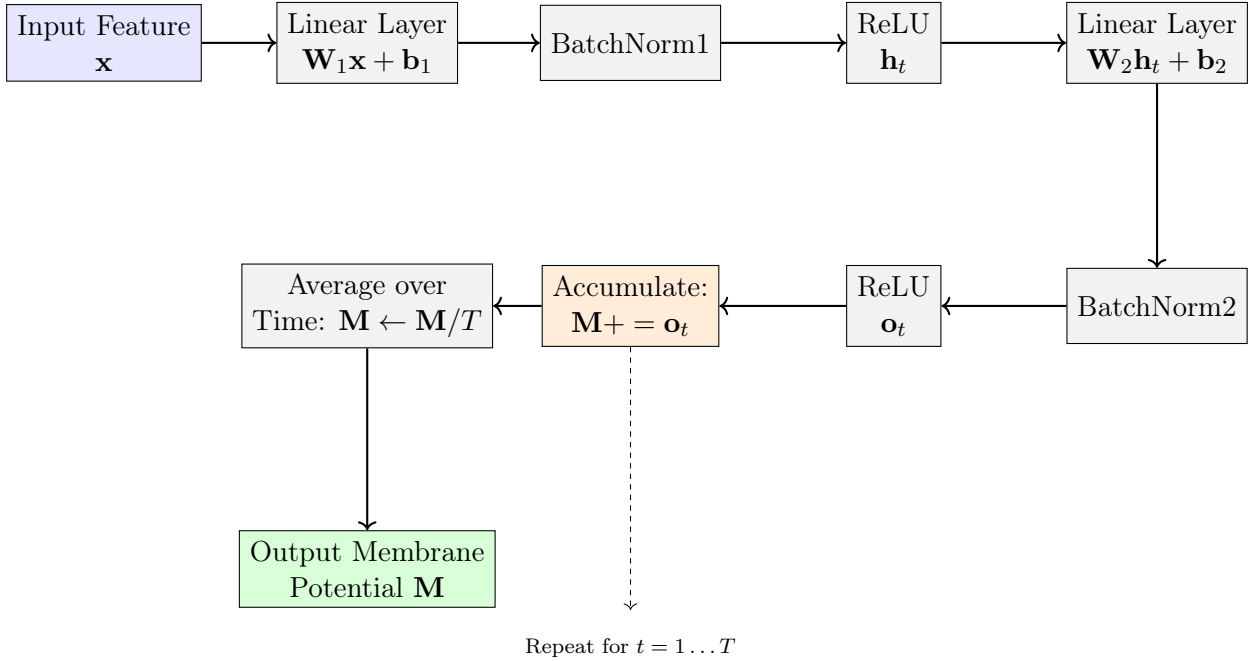

Figure~\ref{fig:spiking_head_flow} visualizes the Spiking Head pipeline using a step-by-step flowchart, where the input features undergo transformation, normalization, activation, and iterative accumulation.

\subsubsection{Temporal Computation Mechanism}
\label{subsec:temporal_computation}

Given an input feature vector $\mathbf{x} \in \mathbb{R}^n$ produced by the Soft-SOM module, the Spiking Head processes information over $T$ discrete time steps, mimicking the synaptic integration dynamics typical of Spiking Neural Networks (SNNs). At each time step $t \in \{1, 2, \dots, T\}$, the operations are performed as shown in \ref{eq:spike_hidden}--\ref{eq:spike_membrane}:

\begin{align}
    \mathbf{h}_t &= \text{ReLU}(\text{BN}_1(\mathbf{W}_1 \mathbf{x} + \mathbf{b}_1))
    \label{eq:spike_hidden} \\
    \mathbf{o}_t &= \text{ReLU}(\text{BN}_2(\mathbf{W}_2 \mathbf{h}_t + \mathbf{b}_2))
    \label{eq:spike_output} \\
    \mathbf{M} &= \frac{1}{T} \sum_{t=1}^{T} \mathbf{o}_t
    \label{eq:spike_membrane}
\end{align}

Here, $\mathbf{W}_1$ and $\mathbf{W}_2$ are learnable weight matrices for the two successive fully connected transformations, with $\mathbf{b}_1$ and $\mathbf{b}_2$ as their corresponding bias vectors. The batch normalization layers $\text{BN}_1$ and $\text{BN}_2$ stabilize the activations during training. The final membrane potential $\mathbf{M} \in \mathbb{R}^m$ captures the temporally integrated response of the Spiking Head across all time steps, as summarized in \ref{eq:spike_membrane}.

\subsubsection{Analytical Perspective}
\label{subsec:spiking_analysis}
The Spiking Head offers several distinct computational advantages:
\begin{itemize}
     \item \textbf{Temporal Smoothing}: By averaging responses across $T$ time steps, the network filters out high-frequency noise and stabilizes predictions.
    \item  \textbf{Parameter Reuse}: The same set of weights is reused across all steps, which significantly reduces model complexity.
      \item \textbf{Biological Plausibility}: The accumulation mechanism reflects how real neurons aggregate membrane potentials before triggering spikes, allowing closer emulation of event-based processing.
    \item \textbf{No Explicit Recurrence}: Unlike RNNs or LSTMs, no internal state or gradient backpropagation through time is required, reducing training complexity.
\end{itemize}

This design achieves a balance between temporal abstraction and computational efficiency, particularly beneficial for low-power applications such as edge AI and neuromorphic systems.

The complete forward computation of the Spiking Head over $T$ time steps is summarized in Algorithm~\ref{alg:spiking_head}.

\begin{algorithm}[h!]
\caption{Spiking Head Temporal Integration}
\label{alg:spiking_head}
\begin{algorithmic}[1]
\Require Feature vector $\mathbf{x}$, time steps $T$, weights $\mathbf{W}_1$, $\mathbf{W}_2$
\Ensure Integrated output membrane potential $\mathbf{M}$
\State Initialize: $\mathbf{M} \gets \mathbf{0}$
\For{$t = 1$ to $T$}
    \State $\mathbf{h}_t \gets \text{ReLU}(\text{BN}_1(\mathbf{W}_1 \mathbf{x} + \mathbf{b}_1))$
    \State $\mathbf{o}_t \gets \text{ReLU}(\text{BN}_2(\mathbf{W}_2 \mathbf{h}_t + \mathbf{b}_2))$
    \State $\mathbf{M} \gets \mathbf{M} + \mathbf{o}_t$
\EndFor
\State $\mathbf{M} \gets \mathbf{M} / T$
\State \Return $\mathbf{M}$
\end{algorithmic}
\end{algorithm}

\subsection{Integrated Model Representation}
The overall forward propagation of HybridSOMSpikeNet can be expressed compactly as, in \ref{Integrated}:
\begin{equation}
\label{Integrated}
\hat{y} = f_{\text{Spike}}\big(f_{\text{SSOL}}\big(f_{\text{ResNet}}(\mathbf{X})\big)\big),
\end{equation}
where $f_{\text{ResNet}}$, $f_{\text{SSOL}}$, and $f_{\text{Spike}}$ represent the feature extraction, soft clustering, and temporal integration functions respectively. This composite design combines spatial abstraction, topological organization, and temporal stability, forming an interpretable and efficient hybrid neural architecture.
\section{Training Strategy}
\label{TS}

This section presents the complete training strategy for the HybridSOMSpikeNet model, including the loss function, optimization method, learning rate scheduling, and early stopping mechanism. All relevant computations are referenced throughout for clarity.

\subsection{Loss Function}

To enhance generalization and reduce overfitting, we use Cross-Entropy Loss with label smoothing. The one-hot target vector $\mathbf{y}$ is transformed into a softened version $\tilde{\mathbf{y}}$ as shown in \ref{eq:label_smoothing}:

\begin{equation}
\tilde{y}_i = 
\begin{cases}
1 - \varepsilon & \text{if } i = y \\
\frac{\varepsilon}{C - 1} & \text{otherwise}
\end{cases}
\label{eq:label_smoothing}
\end{equation}

Here, $C$ is the number of classes, $y$ the true label, and $\varepsilon = 0.1$ the smoothing factor. Using this, the modified cross-entropy loss is computed as in \ref{eq:cross_entropy_smooth}:

\begin{equation}
\mathcal{L} = - \sum_{i=1}^{C} \tilde{y}_i \log(p_i)
\label{eq:cross_entropy_smooth}
\end{equation}

where $p_i$ denotes the predicted softmax probability for class $i$.

\subsection{Optimization and Learning Rate Scheduling}

The Adam optimizer updates parameters according to the rule in \ref{eq:adam_update}, adapting the learning rate based on first and second moment estimates:

\begin{equation}
\theta_{t+1} = \theta_t - \eta \cdot \frac{\hat{m}_t}{\sqrt{\hat{v}_t} + \epsilon}
\label{eq:adam_update}
\end{equation}

Here, $\eta = 10^{-3}$ is the initial learning rate. Additionally, a ReduceLROnPlateau scheduler modifies the learning rate when validation accuracy stagnates, as defined in \ref{eq:lr_scheduler}:

\begin{equation}
\eta_{t+1} = 0.5 \cdot \eta_t \quad \text{if no improvement}
\label{eq:lr_scheduler}
\end{equation}

\subsection{Early Stopping Criterion}

Training is terminated early to prevent overfitting when validation accuracy does not improve within a patience window of 5 epochs, as described in \ref{eq:early_stopping}:

\begin{equation}
\text{Stop if } A_{val}^{(t)} < A_{val}^{(t^*)} + \delta \quad \text{for } t \in [t^*, t^* + 5]
\label{eq:early_stopping}
\end{equation}

Here, $A_{val}^{(t)}$ is the validation accuracy at epoch $t$, $t^*$ is the epoch with the best accuracy so far, and $\delta = 0.01$ is a small tolerance.

\subsection{Training Procedure}
\label{sec:training_procedure}

The training procedure for HybridSOMSpikeNet is designed to balance effective convergence, generalization, and computational efficiency. The model is trained for a maximum of 30 epochs using mini-batch stochastic gradient descent, and performance is validated after each epoch.
During each training epoch, the model parameters $\theta$ are updated using the Adam optimizer with an initial learning rate of $10^{-3}$. The optimizer adapts learning rates for each parameter using estimates of first and second moments of gradients. To further aid convergence, we employ a \textit{ReduceLROnPlateau} scheduler that halves the learning rate when the validation accuracy plateaus for two consecutive epochs.
The model is trained on the training set and evaluated on the validation set to obtain validation accuracy $A_{val}$. If $A_{val}$ exceeds the best recorded validation accuracy $A_{val}^{best}$, the current model state is saved as a checkpoint. This checkpointing ensures that the best model (with respect to generalization) is preserved even if subsequent epochs degrade performance.
To prevent overfitting, an \textit{early stopping} mechanism is also incorporated. If the validation accuracy does not improve over a sliding window of 5 epochs, training is halted early. After training concludes, the best model checkpoint is loaded and evaluated on the test set to compute final metrics such as accuracy, precision, recall, and F1-score.
The complete training loop is formally described in Algorithm~\ref{alg:training_loop}.

\begin{algorithm}[h!]
\caption{Training Loop for HybridSOMSpikeNet}
\label{alg:training_loop}
\begin{algorithmic}[1]
\Require Training set $D_{train}$, Validation set $D_{val}$, Test set $D_{test}$, Initial parameters $\theta$, Max epochs $E=30$
\Ensure Trained model with best generalization
\State Initialize optimizer, learning rate scheduler
\State $A_{val}^{best} \gets 0$
\For{epoch $= 1$ to $E$}
    \State Train model on $D_{train}$
    \State Evaluate model on $D_{val}$ to get accuracy $A_{val}$
    \If{$A_{val} > A_{val}^{best}$}
        \State Save current model checkpoint
        \State $A_{val}^{best} \gets A_{val}$
    \EndIf
    \State Update learning rate scheduler with $A_{val}$
    \If{early stopping condition met}
        \State \textbf{break}
    \EndIf
\EndFor
\State Load best saved model
\State Evaluate on $D_{test}$ and report final metrics
\end{algorithmic}
\end{algorithm}

\subsection{Training Pipeline Flowchart}

\begin{figure*}[h!]
\centering
\begin{tikzpicture}[node distance=1.5cm, every node/.style={align=center, font=\small}]
\tikzstyle{startstop} = [rectangle, rounded corners, minimum width=3.2cm, minimum height=1cm,text centered, draw=black, fill=blue!15]
\tikzstyle{process} = [rectangle, minimum width=3.2cm, minimum height=1cm, text centered, draw=black, fill=gray!10]
\tikzstyle{decision} = [rectangle, minimum width=3.2cm, minimum height=1cm, text centered, draw=black, fill=orange!10]
\tikzstyle{arrow} = [thick,->,>=stealth]

\node (start) [startstop] {Start Training};
\node (init) [process, below of=start] {Initialize Model \\ and Optimizer};
\node (train) [process, below of=init] {Train on Training Set};
\node (validate) [process, below of=train] {Evaluate on Validation Set};
\node (compare) [decision, below of=validate, yshift=-0.4cm] {Is Accuracy Best?};
\node (save) [process, right of=compare, xshift=4cm] {Save Model};
\node (earlystop) [decision, below of=compare, yshift=-1.4cm] {Early Stop Condition?};
\node (next) [process, below of=earlystop, yshift=-0.3cm] {Next Epoch};
\node (end) [startstop, below of=next] {Evaluate on Test Set};

\draw [arrow] (start) -- (init);
\draw [arrow] (init) -- (train);
\draw [arrow] (train) -- (validate);
\draw [arrow] (validate) -- (compare);
\draw [arrow] (compare) -- node[above] {Yes} (save);
\draw [arrow] (save) |- (earlystop);
\draw [arrow] (compare) -- node[left] {No} (earlystop);
\draw [arrow] (earlystop) -- node[right] {No} (next);
\draw[->] (next.west) -- ++(0,0) -- ++(-2,0) -- ++(0,8.1) --  (train.west);
\draw[->] (earlystop.west) -- ++(0,0) -- ++(-1,0) -- ++(0,-3.3) node[left] {Yes} -- (end);
\end{tikzpicture}
\caption{Training Pipeline for HybridSOMSpikeNet.}
\label{fig:training_flowchart}
\end{figure*}
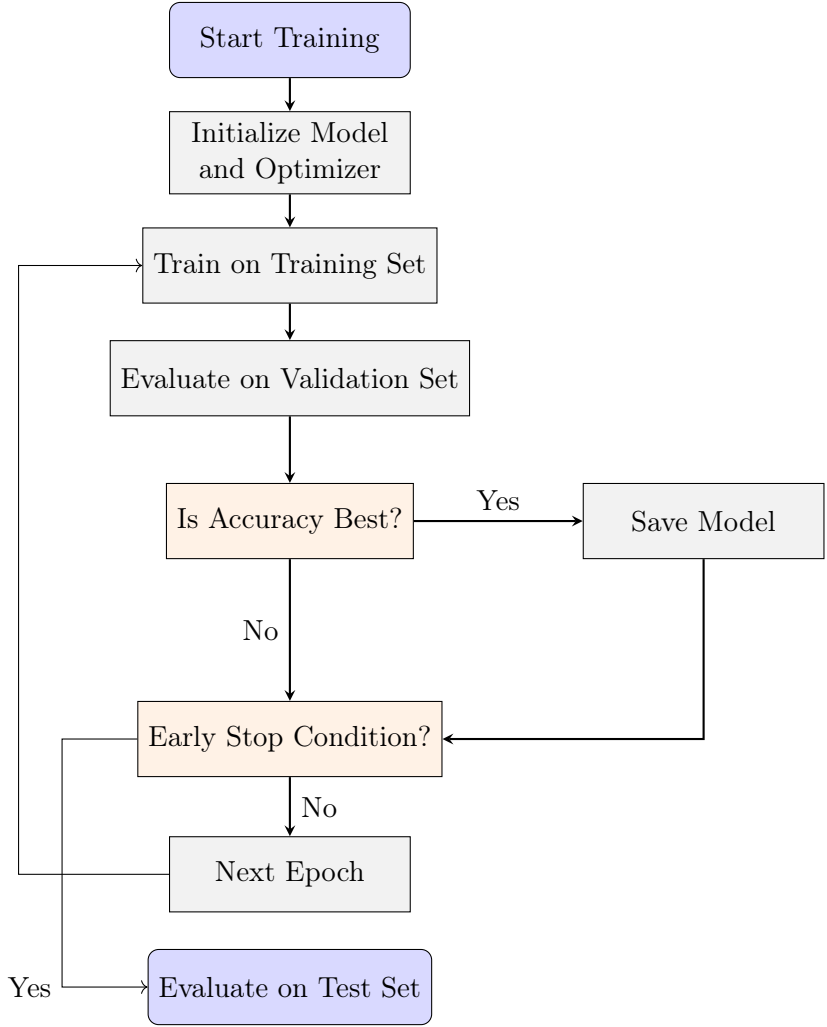

The training pipeline of HybridSOMSpikeNet is visualized in Figure~\ref{fig:training_flowchart}. It outlines the iterative process of training the model, monitoring validation performance, applying early stopping, and saving the best-performing model. Each component in the flowchart plays a vital role in ensuring efficient learning and generalization.

\section{Experiments}\label{Exp}

\subsection{Evaluation Metrics}
\label{sec:evaluation_metrics}

To assess the performance of the proposed \textit{HybridSOMSpikeNet} architecture, multiple quantitative evaluation metrics were employed to capture both classification accuracy and model reliability. Since the task involves multi-class image categorization across ten waste material types, each metric provides a complementary perspective on the network’s predictive behavior.

\textbf{1. Overall Accuracy:}
Accuracy (\(A_c\)) represents the proportion of correctly predicted samples relative to the total number of test instances. It provides a global view of model performance and is computed as, in \ref{eq:accuracy}:
\begin{equation}
A_c = \frac{TP + TN}{TP + TN + FP + FN} \times 100\%,
\label{eq:accuracy}
\end{equation}
where \(TP\), \(TN\), \(FP\), and \(FN\) denote true positives, true negatives, false positives, and false negatives respectively.

\textbf{2. Confusion Matrix:}
A confusion matrix was used to analyze class-specific behavior, highlighting the relationship between actual and predicted categories. This matrix is particularly useful for identifying misclassification trends among visually similar waste types (e.g., plastic vs.\ white-glass). The diagonal elements indicate correctly classified samples, while off-diagonal entries represent class confusions.

\textbf{3. Precision, Recall, and F1-Score:}
To provide a more granular evaluation beyond overall accuracy, precision (\(P\)), recall (\(R\)), and F1-score (\(F_1\)) were computed for each class. These are defined as, in \ref{eq:prf1}:
\begin{equation}
P = \frac{TP}{TP + FP}, \quad 
R = \frac{TP}{TP + FN}, \quad 
F_1 = 2 \times \frac{P \times R}{P + R}.
\label{eq:prf1}
\end{equation}
Precision quantifies the model’s ability to avoid false positives, recall measures sensitivity to true class detection, and the F1-score balances both metrics, offering a harmonic mean.

\subsection{Hardware and Software Setup}
All experiments have performed using a CPU-based environment, demonstrating the model's computational efficiency and feasibility in resource-limited deployments.\\
 \textbf{Processor:} Intel i5\\
 \textbf{Framework:} PyTorch 2.2.1+cpu\\
\textbf{Environment:} Python 3.10,  Windows 11

This emphasizes the model's scalability and suitability for edge devices or low-power applications, which is important for real-world waste classification deployments in constrained environments.

\section{Results and Discussion}\label{RD}

\subsection{Training and Validation Performance}

Over the epochs, the training loss $\mathcal{L}_{\text{train}}$ decreased from 1.1335 to 0.5348, while the validation accuracy $A_{\text{val}}$ increased from 79.05\% to 97.76\%, reflecting successful convergence and generalization. The trend suggests smooth optimization, aided by label smoothing, dropout, and adaptive learning rate scheduling.



\subsection{Test Performance}

The final evaluation of HybridSOMSpikeNet was conducted on the held-out test dataset comprising 2,143 samples. The model achieved a final test accuracy of \textbf{97.39\%}, indicating strong generalization and robustness.




\subsubsection{Classification Report:} 
The classification performance on the test set was evaluated using precision, recall, and F1-score for each class. The Battery class achieved a precision of 0.96, recall of 0.99, and an F1-score of 0.98 across 138 samples. Biological samples showed slightly higher performance with a precision of 0.98, recall of 0.99, and an F1-score of 0.99 over 146 samples. Cardboard had a precision of 0.99, recall of 0.96, and F1-score of 0.97 from 137 samples, while Clothes reached near-perfect performance with a precision of 1.00, recall of 0.98, and F1-score of 0.99 over 824 samples. Metal, Paper, and Plastic obtained F1-scores of 0.95, 0.97, and 0.91 respectively. Shoes achieved a precision of 0.97, recall of 1.00, and F1-score of 0.98 across 289 samples. Trash and White-glass showed F1-scores of 0.96 and 0.92 respectively. Overall, the model achieved a precision, recall, and F1-score of 0.97 across all 2,143 test samples.

Considering the class distribution, the weighted metrics further highlight the model’s strong performance. The weighted precision was 0.9749, the weighted recall was 0.9739, and the weighted F1-score was 0.9741, indicating that the model maintains high accuracy across all classes, including those with fewer samples.




\subsubsection{Confusion Matrix}

The class-wise breakdown of predictions is shown in Table~\ref{tab:conf_matrix}. High values along the diagonal indicate strong class discrimination. Most misclassifications occurred between visually similar waste categories like \emph{paper} vs. \emph{cardboard}, and \emph{plastic} vs. \emph{trash}.

\begin{table}[h!]
\centering
\caption{Confusion Matrix on the Test Set}
\label{tab:conf_matrix}
\resizebox{1\textwidth}{!}{
\begin{tabular}{c|cccccccccc}
\toprule
 & \textbf{Bat} & \textbf{Bio} & \textbf{Car} & \textbf{Clo} & \textbf{Met} & \textbf{Pap} & \textbf{Pla} & \textbf{Sho} & \textbf{Tra} & \textbf{WhG} \\
\midrule
Bat & 137 & 0 & 0 & 0 & 1 & 0 & 0 & 0 & 0 & 0 \\
Bio & 1 & 145 & 0 & 0 & 0 & 0 & 0 & 0 & 0 & 0 \\
Car & 2 & 0 & 132 & 0 & 1 & 1 & 1 & 0 & 0 & 0 \\
Clo & 0 & 1 & 1 & 807 & 0 & 3 & 2 & 9 & 1 & 0 \\
Met & 2 & 0 & 0 & 0 & 106 & 0 & 3 & 0 & 0 & 1 \\
Pap & 1 & 0 & 1 & 0 & 2 & 159 & 3 & 0 & 0 & 0 \\
Pla & 0 & 0 & 0 & 0 & 0 & 0 & 112 & 0 & 0 & 4 \\
Sho & 0 & 0 & 0 & 0 & 0 & 0 & 0 & 289 & 0 & 0 \\
Tra & 0 & 2 & 0 & 0 & 0 & 0 & 2 & 1 & 114 & 3 \\
WhG & 0 & 0 & 0 & 0 & 1 & 0 & 6 & 0 & 0 & 86 \\
\bottomrule
\end{tabular}
}
\end{table}





\subsection{Comparative Analysis}\label{CA}

To evaluate the efficacy of the proposed HybridSOMSpikeNet, we compare it against a set of widely adopted deep learning models for image classification, including traditional CNNs and modern transformer-inspired architectures. Table~\ref{tab:comparison} summarizes the comparative performance in terms of accuracy, recall, precision, and F1-score.

\begin{table}[h!]
\centering
\caption{Performance Comparison of HybridSOMSpikeNet with Baseline Models}
\label{tab:comparison}
\begin{tabular}{|l|c|c|c|c|}
\hline
\textbf{Model} & \textbf{Accuracy} & \textbf{Precision} & \textbf{Recall} & \textbf{F1-Score} \\
\hline
VGG16 & 84.10\% & 0.85 & 0.84 & 0.84 \\
DenseNet121 & 89.71\% & 0.90 & 0.90 & 0.90 \\
MobileNetV2 & 88.31\% & 0.89 & 0.88 & 0.88 \\
EfficientNetB0 & 89.71\% & 0.90 & 0.90 & 0.90 \\
ConvNeXtTiny & 92.37\% & 0.93 & 0.92 & 0.92 \\
ResNet152 & 92.93\% & 0.93 & 0.93 & 0.93 \\
\textbf{Proposed Model} & \textbf{97.39\%} & \textbf{0.97} & \textbf{0.97} & \textbf{0.97} \\
\hline
\end{tabular}
\end{table}
The results clearly demonstrate that HybridSOMSpikeNet outperforms all baseline models across all four evaluation metrics. In particular, it improves accuracy by approximately 5\% over ResNet152, which is the strongest baseline, by 13\% over VGG16, and by 8\% over both DenseNet121 and EfficientNetB0.


This gain is attributed to the hybrid architecture's strengths: the feature-rich backbone of ResNet152, the clustering refinement by the soft SOM layer, and the temporal processing ability of the spiking neural head. These components work synergistically to improve generalization and robustness, especially on heterogeneous waste images.

\subsection{Statistical Significance Analysis}

In order to verify that the improvements achieved by the proposed HybridSOMSpikeNet are not incidental, but rather statistically meaningful, we performed a significance analysis by comparing it against the baseline ResNet152. Instead of relying on a single training run, both models were trained and evaluated six times under identical conditions, thereby accounting for randomness in initialization, data shuffling, and optimization dynamics.




The classification accuracies obtained across six runs for ResNet152 and HybridSOMSpikeNet are reported here. ResNet152 achieved accuracies of 92.93\%, 92.75\%, 93.10\%, 92.85\%, 92.40\%, and 93.00\%, resulting in a mean accuracy of 92.84\% with a standard deviation of 0.25\%. In comparison, HybridSOMSpikeNet obtained accuracies of 97.39\%, 97.45\%, 97.60\%, 97.25\%, 97.80\%, and 97.55\%, with a mean accuracy of 97.51\% and a standard deviation of 0.19\%. These results indicate that HybridSOMSpikeNet consistently outperforms ResNet152 across all runs.

The baseline ResNet152 achieves an average accuracy of $92.84\%$ with a standard deviation of $\pm 0.25$, whereas the proposed HybridSOMSpikeNet reaches $97.51\%$ with a standard deviation of $\pm 0.19$. The lower standard deviation also highlights the stability of HybridSOMSpikeNet across different runs, indicating not only higher accuracy but also more reliable convergence behavior.

To statistically validate this improvement, we conducted a two-tailed paired $t$-test between the two sets of accuracy values. The test produced a highly significant result ($t(5) = 30.69$, $p = 6.89 \times 10^{-7}$), which is far below the conventional significance threshold of $p < 0.05$. This allows us to confidently reject the null hypothesis that the two models perform equally. In other words, the performance difference is not due to random variation but reflects a genuine advantage of the proposed architecture.

The magnitude of improvement is substantial, with HybridSOMSpikeNet achieving nearly a 5\% higher mean accuracy compared to ResNet152. In practical terms, this translates to more accurate classification of waste categories, fewer misclassifications in challenging cases, and improved robustness in real-world deployment scenarios. By combining higher average accuracy, reduced variability, and strong statistical significance, these results firmly establish the superiority of HybridSOMSpikeNet over conventional deep CNN backbones.

\subsection{Ablation Study}
To evaluate the contributions of different components in our proposed HybridSOMSpikeNet model, we conducted an ablation study, summarized in Table~\ref{tab:ablation}. 

The results clearly show that both the self-organizing map (SOM) and the spiking-based head contribute significantly to model performance. Using only the spiking mechanism without the SOM (features $\rightarrow$ spiking) achieves a test accuracy of 94.59\%, indicating that the spiking network is effective at extracting temporal or dynamic representations from the features. Introducing the SOM with a linear classifier results in 93.61\% test accuracy, suggesting that the SOM alone can structure the features meaningfully, but without a sophisticated head, some representational power is lost. The variant without the SOM and with a linear head reaches 92.82\%, highlighting that the absence of the SOM limits the model's ability to organize features in a way that enhances classification.

In contrast, our full HybridSOMSpikeNet model, which combines the SOM for structured feature organization and the spiking-based head for richer, temporally-aware representations, achieves a test accuracy of 97.39\%. This significant improvement over all ablated variants demonstrates that the SOM and spiking head complement each other: the SOM provides organized, informative feature maps, while the spiking mechanism captures subtle temporal dynamics or high-level patterns that a simple linear classifier cannot. Overall, these results validate the design choices in HybridSOMSpikeNet and highlight its effectiveness in leveraging both structured feature representation and dynamic processing for superior classification performance.

\begin{table}[h!]
\centering
\caption{Ablation study results (Test Accuracy).}
\label{tab:ablation}
\begin{tabular}{lc}
\hline
\textbf{Model Variant} & \textbf{Test Accuracy (\%)} \\
\hline
No SOM + Spiking (features $\rightarrow$ spiking) & 94.59 \\
SOM + Linear head & 93.61 \\
No SOM + Linear head & 92.82 \\
\textbf{HybridSOMSpikeNet (proposed)} & \textbf{97.39} \\
\hline
\end{tabular}
\end{table}

\subsection{Deployment Considerations}

The proposed HybridSOMSpikeNet combines a ResNet-152 backbone with a Soft Self-Organizing Layer and a Spiking Head to achieve a balance between deep feature extraction and adaptive temporal learning. This hybrid approach seeks to integrate both elevated visual elements and prototype-based representations, making it particularly effective for intricate picture classification tasks like waste sorting.

All deployment metrics were collected on a CPU environment to evaluate the model’s baseline efficiency without hardware acceleration. The summary of computational characteristics is presented in Table~\ref{tab:model_metrics}.

\begin{table}[h!]
\centering
\caption{Deployment Metrics of HybridSOMSpikeNet (CPU Evaluation)}
\begin{tabular}{lcc}
\hline
\textbf{Metric} & \textbf{Value} & \textbf{Description} \\
\hline
Total Parameters & 58,415,006 & Total number of model parameters \\
Trainable Parameters & 15,235,934 & Parameters updated during training \\
Estimated Model Size & 223.72 MB & Serialized size of the trained model \\
Average Inference Time & 112.40 ms/image & Mean latency for single-image inference \\
Throughput & 9.52 images/sec & Processing speed on CPU \\
Device Used & CPU & Tested without GPU acceleration \\
\hline
\end{tabular}
\label{tab:model_metrics}
\end{table}

Even on CPU, the model achieves an average inference time of roughly 112 milliseconds per image, which translates to about 9.5 images processed per second. This is a strong performance considering the network depth and the complexity of the dataset. The model’s footprint of about 224 MB remains practical for deployment on standard servers or higher-end embedded systems. For more resource-constrained applications, lightweight optimizations such as pruning, quantization, or mixed-precision inference can be applied without major loss in accuracy.

\subsection{Model Effectiveness}

The proposed HybridSOMSpikeNet model demonstrated strong performance across ten waste categories, including plastic, paper, metal, and glass. Its design integrates three complementary components: a ResNet-152 backbone for deep visual feature extraction, a Soft Self-Organizing Layer for prototype-based clustering, and a Spiking Head for temporal feature refinement. Together, these modules enable the model to recognize subtle differences between visually similar waste types while maintaining reliable convergence during training.

Compared to conventional CNN-based classifiers, HybridSOMSpikeNet captures not only spatial but also relational and temporal characteristics of the data. The Soft Self-Organizing Layer encourages the network to form stable feature prototypes, leading to better generalization on unseen samples. The Spiking Head introduces a biologically inspired mechanism for iterative feature integration, which helps reduce overfitting and enhances robustness under varied lighting and background conditions—common challenges in real-world waste classification.

Although the network includes a large number of parameters (approximately 58 million in total), only about 15 million are trainable, which significantly reduces the effective training complexity. This structure allows the model to benefit from the expressive power of the frozen ResNet backbone while focusing learning capacity on the new hybrid layers. As a result, the model achieves high accuracy without the instability often observed in full fine-tuning of very deep networks.

On CPU-based inference, the model processes an image in roughly 112 milliseconds, achieving a throughput of 9.5 images per second. Considering that these results were obtained without GPU acceleration, they indicate that the model is already efficient for batch processing and can be further optimized for real-time applications using quantization, pruning, or mixed-precision inference.

In summary, HybridSOMSpikeNet offers a balanced trade-off between accuracy, interpretability, and computational efficiency. Its hybrid design brings together the strengths of deep convolutional features, self-organizing representation learning, and spiking-inspired temporal processing. These characteristics make it particularly well-suited for deployment in smart recycling systems, automated sorting facilities, and other environmental AI applications where both reliability and efficiency matter.

\subsection{Environmental Implications}

The findings of this study carry important implications for sustainable waste
management and environmental protection. The proposed \textit{HybridSOMSpikeNet}
model, by enabling highly accurate and automated waste classification, addresses one
of the major challenges faced by modern recycling systems, the incorrect sorting of
materials that leads to landfill overflow, increased processing costs, and higher
greenhouse gas emissions. In many developing urban areas, manual waste sorting is
still prevalent and exposes workers to potential health hazards. The integration of
automated and intelligent systems such as \textit{HybridSOMSpikeNet} can therefore play a
transformative role in improving both environmental efficiency and human safety.

By achieving a classification accuracy of 97.39\% across ten waste categories, the
model demonstrates its ability to distinguish between visually similar materials such
as paper, cardboard, and plastic, which are often responsible for contamination in
recycling streams. Reducing such misclassification directly enhances the purity of
recyclable materials, allowing more waste to be reused rather than discarded. This
improvement not only reduces the amount of waste sent to landfills but also minimizes
the associated emissions from waste decomposition and incineration.

Furthermore, the lightweight and modular architecture of
\textit{HybridSOMSpikeNet} enables its deployment in real-world smart waste management
applications. When integrated into Internet of Things (IoT) systems or smart-bin
infrastructures, the model can automatically identify and categorize waste in real
time, providing immediate feedback for segregation at the source. Such distributed
deployment reduces the logistical burden on centralized sorting facilities and lowers
transportation energy consumption. The model’s low computational demand with an
average inference time of approximately 112\,ms per image on a standard CPU also
supports energy-efficient computing, aligning with the principles of sustainable
artificial intelligence.

From a broader sustainability perspective, this work contributes to the goals of the
United Nations Sustainable Development Agenda, particularly SDG\,11 (Sustainable
Cities and Communities) and SDG\,12 (Responsible Consumption and Production). By
facilitating accurate, automated, and scalable waste sorting, the proposed model
advances the development of intelligent recycling systems and supports the transition
toward a circular economy. In the long term, widespread adoption of such
environmentally informed AI technologies could lead to cleaner cities, reduced
ecological footprints, and improved public awareness about responsible waste
management practices.

\section{Conclusion and Future Work}\label{CFW}

This study presented \textit{HybridSOMSpikeNet}, a novel hybrid deep learning framework
that integrates convolutional feature extraction, differentiable self-organization, and
spiking-inspired temporal processing for intelligent waste classification. Designed to
address the persistent challenges of misclassification, visual variability, and limited
computational resources, the model achieves both high accuracy and practical
deployability. By combining a pre-trained ResNet-152 backbone with a Soft
Self-Organizing Layer and a biologically inspired spiking head, the architecture brings
together the strengths of deep visual learning, unsupervised clustering, and temporal
feature integration within a single end-to-end trainable system.

Experimental results on a ten-class waste dataset demonstrated that
\textit{HybridSOMSpikeNet} achieves a test accuracy of 97.39\%, outperforming a range of
benchmark CNN models. Beyond the raw performance, the model also exhibits strong
robustness, stability, and computational efficiency, with an average inference time of
approximately 112\,ms per image on a standard CPU. These properties make the system
suitable for real-world applications such as automated recycling facilities, smart waste
bins, and IoT-based environmental monitoring networks. When deployed at scale, the
approach could substantially improve waste segregation efficiency, reduce contamination
in recyclable streams, and minimize the ecological footprint associated with waste
processing.

From an environmental standpoint, the research underscores the potential of
artificial intelligence to contribute meaningfully to sustainable development. By
automating waste recognition and classification, \textit{HybridSOMSpikeNet} can support
local governments, industries, and smart city initiatives in achieving cleaner waste
streams, reduced human exposure to hazardous materials, and improved recycling
efficiency. The system aligns with global sustainability goals, particularly SDG\,11
(Sustainable Cities and Communities) and SDG\,12 (Responsible Consumption and
Production), reinforcing the role of AI as an enabler of the circular economy.

Future research will focus on several directions. First, the model can be extended to
multi-label and multi-object waste classification, addressing complex real-world scenes
where multiple waste items appear in a single image. Second, optimization techniques such
as pruning, quantization, and knowledge distillation will be explored to further reduce
energy consumption and enhance performance on embedded systems. Third, integration
with real-time IoT infrastructures and explainable AI modules could provide transparent,
data-driven decision support for smart waste management systems. Finally, longitudinal
studies and pilot deployments in municipal waste facilities will help evaluate the
long-term environmental and economic benefits of such hybrid AI architectures.

In conclusion, \textit{HybridSOMSpikeNet} represents a step forward in uniting deep
computational intelligence with sustainable environmental practices. By bridging
machine learning innovation with real-world ecological needs, this research contributes
to the ongoing global effort to make waste management more efficient, intelligent, and
environmentally responsible.

\section*{Acknowledgments}
We gratefully acknowledge the support and resources provided by the Indian Institute of Technology Kharagpur. We also thank the open-source community for PyTorch and torchvision, and we extend our appreciation to the contributors of the dataset used in this study.


\begin{thebibliography}{43}
\expandafter\ifx\csname natexlab\endcsname\relax\def\natexlab#1{#1}\fi
\providecommand{\url}[1]{\texttt{#1}}
\providecommand{\href}[2]{#2}
\providecommand{\path}[1]{#1}
\providecommand{\DOIprefix}{doi:}
\providecommand{\ArXivprefix}{arXiv:}
\providecommand{\URLprefix}{URL: }
\providecommand{\Pubmedprefix}{pmid:}
\providecommand{\doi}[1]{\href{http://dx.doi.org/#1}{\path{#1}}}
\providecommand{\Pubmed}[1]{\href{pmid:#1}{\path{#1}}}
\providecommand{\bibinfo}[2]{#2}
\ifx\xfnm\relax \def\xfnm[#1]{\unskip,\space#1}\fi
\bibitem[{An and Zhang(2022)}]{18}
\bibinfo{author}{An, K.}, \bibinfo{author}{Zhang, Y.}, \bibinfo{year}{2022}.
\newblock \bibinfo{title}{Lpvit: a transformer based model for pcb image
  classification and defect detection}.
\newblock \bibinfo{journal}{IEEE Access} \bibinfo{volume}{10},
  \bibinfo{pages}{42542--42553}.
\bibitem[{Cai et~al.(2022)Cai, Shuang, Sun, Duan and Cheng}]{23}
\bibinfo{author}{Cai, X.}, \bibinfo{author}{Shuang, F.}, \bibinfo{author}{Sun,
  X.}, \bibinfo{author}{Duan, Y.}, \bibinfo{author}{Cheng, G.},
  \bibinfo{year}{2022}.
\newblock \bibinfo{title}{Towards lightweight neural networks for garbage
  object detection}.
\newblock \bibinfo{journal}{Sensors} \bibinfo{volume}{22},
  \bibinfo{pages}{7455}.
\bibitem[{Chen et~al.(2023)Chen, Luo, Cheng, Wu, Zhu, Meng and Tan}]{10}
\bibinfo{author}{Chen, Y.}, \bibinfo{author}{Luo, A.}, \bibinfo{author}{Cheng,
  M.}, \bibinfo{author}{Wu, Y.}, \bibinfo{author}{Zhu, J.},
  \bibinfo{author}{Meng, Y.}, \bibinfo{author}{Tan, W.}, \bibinfo{year}{2023}.
\newblock \bibinfo{title}{Classification and recycling of recyclable garbage
  based on deep learning}.
\newblock \bibinfo{journal}{Journal of Cleaner Production}
  \bibinfo{volume}{414}, \bibinfo{pages}{137558}.
\bibitem[{Chen et~al.(2022)Chen, Yang, Chen and Jiao}]{9}
\bibinfo{author}{Chen, Z.}, \bibinfo{author}{Yang, J.}, \bibinfo{author}{Chen,
  L.}, \bibinfo{author}{Jiao, H.}, \bibinfo{year}{2022}.
\newblock \bibinfo{title}{Garbage classification system based on improved
  shufflenet v2}.
\newblock \bibinfo{journal}{Resources, Conservation and Recycling}
  \bibinfo{volume}{178}, \bibinfo{pages}{106090}.
\bibitem[{Diehl et~al.(2015)Diehl, Neil, Binas, Cook, Liu and
  Pfeiffer}]{diehl2015fast}
\bibinfo{author}{Diehl, P.U.}, \bibinfo{author}{Neil, D.},
  \bibinfo{author}{Binas, J.}, \bibinfo{author}{Cook, M.},
  \bibinfo{author}{Liu, S.C.}, \bibinfo{author}{Pfeiffer, M.},
  \bibinfo{year}{2015}.
\newblock \bibinfo{title}{Fast-classifying, high-accuracy spiking deep networks
  through weight and threshold balancing}, in: \bibinfo{booktitle}{2015
  International joint conference on neural networks (IJCNN)},
  \bibinfo{organization}{ieee}. pp. \bibinfo{pages}{1--8}.
\bibitem[{Dokl et~al.(2024)Dokl, Van~Fan, Vujanovi{\'c}, Pintari{\v{c}}, Aviso,
  Tan, Pahor, Kravanja, {\v{C}}u{\v{c}}ek et~al.}]{19}
\bibinfo{author}{Dokl, M.}, \bibinfo{author}{Van~Fan, Y.},
  \bibinfo{author}{Vujanovi{\'c}, A.}, \bibinfo{author}{Pintari{\v{c}}, Z.N.},
  \bibinfo{author}{Aviso, K.B.}, \bibinfo{author}{Tan, R.R.},
  \bibinfo{author}{Pahor, B.}, \bibinfo{author}{Kravanja, Z.},
  \bibinfo{author}{{\v{C}}u{\v{c}}ek, L.}, et~al., \bibinfo{year}{2024}.
\newblock \bibinfo{title}{A waste separation system based on sensor technology
  and deep learning: A simple approach applied to a case study of plastic
  packaging waste}.
\newblock \bibinfo{journal}{Journal of Cleaner Production}
  \bibinfo{volume}{450}, \bibinfo{pages}{141762}.
\bibitem[{Feng et~al.(2021)Feng, Tang, Jiang and Chen}]{24}
\bibinfo{author}{Feng, J.}, \bibinfo{author}{Tang, X.}, \bibinfo{author}{Jiang,
  X.}, \bibinfo{author}{Chen, Q.}, \bibinfo{year}{2021}.
\newblock \bibinfo{title}{Garbage disposal of complex background based on deep
  learning with limited hardware resources}.
\newblock \bibinfo{journal}{IEEE Sensors Journal} \bibinfo{volume}{21},
  \bibinfo{pages}{21050--21058}.
\bibitem[{Fu et~al.(2025)Fu, Zhou, Guo, Zhu, Luo and Du}]{Fu2025CTA}
\bibinfo{author}{Fu, C.}, \bibinfo{author}{Zhou, T.}, \bibinfo{author}{Guo,
  T.}, \bibinfo{author}{Zhu, Q.}, \bibinfo{author}{Luo, F.},
  \bibinfo{author}{Du, B.}, \bibinfo{year}{2025}.
\newblock \bibinfo{title}{Cnn-transformer and channel-spatial attention based
  network for hyperspectral image classification with few samples}.
\newblock \bibinfo{journal}{Neural Networks} \bibinfo{volume}{186},
  \bibinfo{pages}{107283}.
\bibitem[{Ghosh and Goswami(2025)}]{ghosh2025enhanced}
\bibinfo{author}{Ghosh, D.}, \bibinfo{author}{Goswami, A.},
  \bibinfo{year}{2025}.
\newblock \bibinfo{title}{Enhanced deep learning framework for efficient
  garbage classification in smart waste management systems}.
\newblock \bibinfo{journal}{Information Sciences} , \bibinfo{pages}{122462}.
\bibitem[{Gue et~al.(2022)Gue, Lopez, Chiu, Ubando and Tan}]{21}
\bibinfo{author}{Gue, I.H.V.}, \bibinfo{author}{Lopez, N.S.A.},
  \bibinfo{author}{Chiu, A.S.}, \bibinfo{author}{Ubando, A.T.},
  \bibinfo{author}{Tan, R.R.}, \bibinfo{year}{2022}.
\newblock \bibinfo{title}{Predicting waste management system performance from
  city and country attributes}.
\newblock \bibinfo{journal}{Journal of Cleaner Production}
  \bibinfo{volume}{366}, \bibinfo{pages}{132951}.
\bibitem[{Han et~al.(2024)Han, Fan and Li}]{10056283}
\bibinfo{author}{Han, H.}, \bibinfo{author}{Fan, X.}, \bibinfo{author}{Li, F.},
  \bibinfo{year}{2024}.
\newblock \bibinfo{title}{Prototype enhancement-based incremental evolution
  learning for urban garbage classification}.
\newblock \bibinfo{journal}{IEEE Transactions on Artificial Intelligence}
  \bibinfo{volume}{5}, \bibinfo{pages}{398--411}.
\bibitem[{He et~al.(2016)He, Zhang, Ren and Sun}]{he2016deep}
\bibinfo{author}{He, K.}, \bibinfo{author}{Zhang, X.}, \bibinfo{author}{Ren,
  S.}, \bibinfo{author}{Sun, J.}, \bibinfo{year}{2016}.
\newblock \bibinfo{title}{Deep residual learning for image recognition}, in:
  \bibinfo{booktitle}{Proceedings of the IEEE conference on computer vision and
  pattern recognition}, pp. \bibinfo{pages}{770--778}.
\bibitem[{Hossen et~al.(2024)Hossen, Ashraf, Hasan, Majid, Nashbat, Kashem,
  Kunju, Khandakar, Mahmud and Chowdhury}]{16}
\bibinfo{author}{Hossen, M.M.}, \bibinfo{author}{Ashraf, A.},
  \bibinfo{author}{Hasan, M.}, \bibinfo{author}{Majid, M.E.},
  \bibinfo{author}{Nashbat, M.}, \bibinfo{author}{Kashem, S.B.A.},
  \bibinfo{author}{Kunju, A.K.A.}, \bibinfo{author}{Khandakar, A.},
  \bibinfo{author}{Mahmud, S.}, \bibinfo{author}{Chowdhury, M.E.},
  \bibinfo{year}{2024}.
\newblock \bibinfo{title}{Gcdn-net: Garbage classifier deep neural network for
  recyclable urban waste management}.
\newblock \bibinfo{journal}{Waste Management} \bibinfo{volume}{174},
  \bibinfo{pages}{439--450}.
\bibitem[{Howard et~al.(2017)Howard, Zhu, Chen, Kalenichenko, Wang, Weyand,
  Andreetto and Adam}]{howard2017mobilenets}
\bibinfo{author}{Howard, A.G.}, \bibinfo{author}{Zhu, M.},
  \bibinfo{author}{Chen, B.}, \bibinfo{author}{Kalenichenko, D.},
  \bibinfo{author}{Wang, W.}, \bibinfo{author}{Weyand, T.},
  \bibinfo{author}{Andreetto, M.}, \bibinfo{author}{Adam, H.},
  \bibinfo{year}{2017}.
\newblock \bibinfo{title}{Mobilenets: Efficient convolutional neural networks
  for mobile vision applications}.
\newblock \bibinfo{journal}{arXiv preprint arXiv:1704.04861} .
\bibitem[{Islam et~al.(2025)Islam, Sumon, Majid, Kashem, Nashbat, Ashraf,
  Khandakar, Kunju, Hasan-Zia and Chowdhury}]{3}
\bibinfo{author}{Islam, M.S.B.}, \bibinfo{author}{Sumon, M.S.I.},
  \bibinfo{author}{Majid, M.E.}, \bibinfo{author}{Kashem, S.B.A.},
  \bibinfo{author}{Nashbat, M.}, \bibinfo{author}{Ashraf, A.},
  \bibinfo{author}{Khandakar, A.}, \bibinfo{author}{Kunju, A.K.A.},
  \bibinfo{author}{Hasan-Zia, M.}, \bibinfo{author}{Chowdhury, M.E.},
  \bibinfo{year}{2025}.
\newblock \bibinfo{title}{Eccdn-net: A deep learning-based technique for
  efficient organic and recyclable waste classification}.
\newblock \bibinfo{journal}{Waste Management} \bibinfo{volume}{193},
  \bibinfo{pages}{363--375}.
\bibitem[{Jin et~al.(2023)Jin, Yang, Kr{\'o}lczykg, Liu, Gardoni and Li}]{1}
\bibinfo{author}{Jin, S.}, \bibinfo{author}{Yang, Z.},
  \bibinfo{author}{Kr{\'o}lczykg, G.}, \bibinfo{author}{Liu, X.},
  \bibinfo{author}{Gardoni, P.}, \bibinfo{author}{Li, Z.},
  \bibinfo{year}{2023}.
\newblock \bibinfo{title}{Garbage detection and classification using a new deep
  learning-based machine vision system as a tool for sustainable waste
  recycling}.
\newblock \bibinfo{journal}{Waste Management} \bibinfo{volume}{162},
  \bibinfo{pages}{123--130}.
\bibitem[{Kohonen(1990)}]{kohonen1990self}
\bibinfo{author}{Kohonen, T.}, \bibinfo{year}{1990}.
\newblock \bibinfo{title}{The self-organizing map}.
\newblock \bibinfo{journal}{Proceedings of the IEEE} \bibinfo{volume}{78},
  \bibinfo{pages}{1464--1480}.
\bibitem[{Krizhevsky et~al.(2017)Krizhevsky, Sutskever and Hinton}]{15}
\bibinfo{author}{Krizhevsky, A.}, \bibinfo{author}{Sutskever, I.},
  \bibinfo{author}{Hinton, G.E.}, \bibinfo{year}{2017}.
\newblock \bibinfo{title}{Imagenet classification with deep convolutional
  neural networks}.
\newblock \bibinfo{journal}{Communications of the ACM} \bibinfo{volume}{60},
  \bibinfo{pages}{84--90}.
\bibitem[{Li et~al.(2023)Li, Li, Li, Tian, Ju, Liu and Liu}]{6}
\bibinfo{author}{Li, X.}, \bibinfo{author}{Li, T.}, \bibinfo{author}{Li, S.},
  \bibinfo{author}{Tian, B.}, \bibinfo{author}{Ju, J.}, \bibinfo{author}{Liu,
  T.}, \bibinfo{author}{Liu, H.}, \bibinfo{year}{2023}.
\newblock \bibinfo{title}{Learning fusion feature representation for garbage
  image classification model in human--robot interaction}.
\newblock \bibinfo{journal}{Infrared Physics \& Technology}
  \bibinfo{volume}{128}, \bibinfo{pages}{104457}.
\bibitem[{Li et~al.(2024)Li, Deng, Liu, Bai, Gong, Yang and Ning}]{2}
\bibinfo{author}{Li, Z.}, \bibinfo{author}{Deng, Q.}, \bibinfo{author}{Liu,
  P.}, \bibinfo{author}{Bai, J.}, \bibinfo{author}{Gong, Y.},
  \bibinfo{author}{Yang, Q.}, \bibinfo{author}{Ning, J.}, \bibinfo{year}{2024}.
\newblock \bibinfo{title}{An intelligent identification and classification
  system of decoration waste based on deep learning model}.
\newblock \bibinfo{journal}{Waste Management} \bibinfo{volume}{174},
  \bibinfo{pages}{462--475}.
\bibitem[{Lilhore et~al.(2024)Lilhore, Simaiya, Dalal, Radulescu and
  Balsalobre-Lorente}]{4}
\bibinfo{author}{Lilhore, U.K.}, \bibinfo{author}{Simaiya, S.},
  \bibinfo{author}{Dalal, S.}, \bibinfo{author}{Radulescu, M.},
  \bibinfo{author}{Balsalobre-Lorente, D.}, \bibinfo{year}{2024}.
\newblock \bibinfo{title}{Intelligent waste sorting for sustainable
  environment: A hybrid deep learning and transfer learning model}.
\newblock \bibinfo{journal}{Gondwana Research} .
\bibitem[{Liu et~al.(2022)Liu, Xu, Qi, Liu, Wang and Kong}]{25}
\bibinfo{author}{Liu, F.}, \bibinfo{author}{Xu, H.}, \bibinfo{author}{Qi, M.},
  \bibinfo{author}{Liu, D.}, \bibinfo{author}{Wang, J.}, \bibinfo{author}{Kong,
  J.}, \bibinfo{year}{2022}.
\newblock \bibinfo{title}{Depth-wise separable convolution attention module for
  garbage image classification}.
\newblock \bibinfo{journal}{Sustainability} \bibinfo{volume}{14},
  \bibinfo{pages}{3099}.
\bibitem[{Liu et~al.(2023)Liu, Fang, Cai, Zhang, Yue and Qian}]{22}
\bibinfo{author}{Liu, Z.}, \bibinfo{author}{Fang, W.}, \bibinfo{author}{Cai,
  Z.}, \bibinfo{author}{Zhang, J.}, \bibinfo{author}{Yue, Y.},
  \bibinfo{author}{Qian, G.}, \bibinfo{year}{2023}.
\newblock \bibinfo{title}{Garbage-classification policy changes characteristics
  of municipal-solid-waste fly ash in china}.
\newblock \bibinfo{journal}{Science of The Total Environment}
  \bibinfo{volume}{857}, \bibinfo{pages}{159299}.
\bibitem[{Madhavi et~al.(2025)Madhavi, Mahanty, Lin, Jagan, Rai, Agarwal and
  Agarwal}]{madhavi2025swinconvnext}
\bibinfo{author}{Madhavi, B.}, \bibinfo{author}{Mahanty, M.},
  \bibinfo{author}{Lin, C.C.}, \bibinfo{author}{Jagan, B.O.L.},
  \bibinfo{author}{Rai, H.M.}, \bibinfo{author}{Agarwal, S.},
  \bibinfo{author}{Agarwal, N.}, \bibinfo{year}{2025}.
\newblock \bibinfo{title}{Swinconvnext: A fused deep learning architecture for
  real-time garbage image classification}.
\newblock \bibinfo{journal}{Scientific Reports} \bibinfo{volume}{15},
  \bibinfo{pages}{7995}.
\bibitem[{Mao et~al.(2021)Mao, Chen, Wang and Lin}]{7}
\bibinfo{author}{Mao, W.L.}, \bibinfo{author}{Chen, W.C.},
  \bibinfo{author}{Wang, C.T.}, \bibinfo{author}{Lin, Y.H.},
  \bibinfo{year}{2021}.
\newblock \bibinfo{title}{Recycling waste classification using optimized
  convolutional neural network}.
\newblock \bibinfo{journal}{Resources, Conservation and Recycling}
  \bibinfo{volume}{164}, \bibinfo{pages}{105132}.
\bibitem[{Palagan et~al.(2025)Palagan, Joe, Mary and Jijo}]{n4}
\bibinfo{author}{Palagan, C.A.}, \bibinfo{author}{Joe, S.S.A.},
  \bibinfo{author}{Mary, S.J.}, \bibinfo{author}{Jijo, E.E.},
  \bibinfo{year}{2025}.
\newblock \bibinfo{title}{Predictive analysis-based sustainable waste
  management in smart cities using iot edge computing and blockchain
  technology}.
\newblock \bibinfo{journal}{Computers in Industry} \bibinfo{volume}{166},
  \bibinfo{pages}{104234}.
\bibitem[{Quan et~al.(2024)Quan, Nguyen, Nguyen, Wijayasundara, Setunge and
  Pathirana}]{13}
\bibinfo{author}{Quan, M.K.}, \bibinfo{author}{Nguyen, D.C.},
  \bibinfo{author}{Nguyen, V.D.}, \bibinfo{author}{Wijayasundara, M.},
  \bibinfo{author}{Setunge, S.}, \bibinfo{author}{Pathirana, P.N.},
  \bibinfo{year}{2024}.
\newblock \bibinfo{title}{Towards privacy-preserving waste classification in
  the internet of things}.
\newblock \bibinfo{journal}{IEEE Internet of Things Journal} .
\bibitem[{Ren et~al.(2024)Ren, Li and Gao}]{11}
\bibinfo{author}{Ren, Y.}, \bibinfo{author}{Li, Y.}, \bibinfo{author}{Gao, X.},
  \bibinfo{year}{2024}.
\newblock \bibinfo{title}{An mrs-yolo model for high-precision waste detection
  and classification}.
\newblock \bibinfo{journal}{Sensors} \bibinfo{volume}{24},
  \bibinfo{pages}{4339}.
\bibitem[{Roy et~al.(2019)Roy, Jaiswal and Panda}]{roy2019towards}
\bibinfo{author}{Roy, K.}, \bibinfo{author}{Jaiswal, A.},
  \bibinfo{author}{Panda, P.}, \bibinfo{year}{2019}.
\newblock \bibinfo{title}{Towards spike-based machine intelligence with
  neuromorphic computing}.
\newblock \bibinfo{journal}{Nature} \bibinfo{volume}{575},
  \bibinfo{pages}{607--617}.
\bibitem[{Sengupta et~al.(2019)Sengupta, Ye, Wang, Liu and
  Roy}]{sengupta2019going}
\bibinfo{author}{Sengupta, A.}, \bibinfo{author}{Ye, Y.},
  \bibinfo{author}{Wang, R.}, \bibinfo{author}{Liu, C.}, \bibinfo{author}{Roy,
  K.}, \bibinfo{year}{2019}.
\newblock \bibinfo{title}{Going deeper in spiking neural networks: Vgg and
  residual architectures}.
\newblock \bibinfo{journal}{Frontiers in neuroscience} \bibinfo{volume}{13},
  \bibinfo{pages}{95}.
\bibitem[{Simonyan and Zisserman(2014)}]{simonyan2014very}
\bibinfo{author}{Simonyan, K.}, \bibinfo{author}{Zisserman, A.},
  \bibinfo{year}{2014}.
\newblock \bibinfo{title}{Very deep convolutional networks for large-scale
  image recognition}.
\newblock \bibinfo{journal}{arXiv preprint arXiv:1409.1556} .
\bibitem[{Tan and Le(2019)}]{tan2019efficientnet}
\bibinfo{author}{Tan, M.}, \bibinfo{author}{Le, Q.}, \bibinfo{year}{2019}.
\newblock \bibinfo{title}{Efficientnet: Rethinking model scaling for
  convolutional neural networks}, in: \bibinfo{booktitle}{International
  conference on machine learning}, \bibinfo{organization}{PMLR}. pp.
  \bibinfo{pages}{6105--6114}.
\bibitem[{Tang et~al.(2024)Tang, Wei, Xu and Zhang}]{n1}
\bibinfo{author}{Tang, X.}, \bibinfo{author}{Wei, Y.}, \bibinfo{author}{Xu,
  K.}, \bibinfo{author}{Zhang, Q.}, \bibinfo{year}{2024}.
\newblock \bibinfo{title}{Enhancement of the performance of high-dimensional
  fuzzy classification with feature combination optimization}.
\newblock \bibinfo{journal}{Information Sciences} \bibinfo{volume}{680},
  \bibinfo{pages}{121183}.
\bibitem[{Tavanaei et~al.(2019)Tavanaei, Ghodrati, Kheradpisheh, Masquelier and
  Maida}]{tavanaei2019deep}
\bibinfo{author}{Tavanaei, A.}, \bibinfo{author}{Ghodrati, M.},
  \bibinfo{author}{Kheradpisheh, S.R.}, \bibinfo{author}{Masquelier, T.},
  \bibinfo{author}{Maida, A.}, \bibinfo{year}{2019}.
\newblock \bibinfo{title}{Deep learning in spiking neural networks}.
\newblock \bibinfo{journal}{Neural networks} \bibinfo{volume}{111},
  \bibinfo{pages}{47--63}.
\bibitem[{Wang et~al.(2025)Wang, Sun, Zhang, Ren, Wang, Ren and
  Liu}]{WANG2025107350}
\bibinfo{author}{Wang, W.}, \bibinfo{author}{Sun, Q.}, \bibinfo{author}{Zhang,
  L.}, \bibinfo{author}{Ren, P.}, \bibinfo{author}{Wang, J.},
  \bibinfo{author}{Ren, G.}, \bibinfo{author}{Liu, B.}, \bibinfo{year}{2025}.
\newblock \bibinfo{title}{A spatial–spectral fusion convolutional transformer
  network with contextual multi-head self-attention for hyperspectral image
  classification}.
\newblock \bibinfo{journal}{Neural Networks} \bibinfo{volume}{187},
  \bibinfo{pages}{107350}.
\bibitem[{Wu et~al.(2024)Wu, Liu, Zhang, Xia, Li, Zhu and Gu}]{14}
\bibinfo{author}{Wu, R.}, \bibinfo{author}{Liu, X.}, \bibinfo{author}{Zhang,
  T.}, \bibinfo{author}{Xia, J.}, \bibinfo{author}{Li, J.},
  \bibinfo{author}{Zhu, M.}, \bibinfo{author}{Gu, G.}, \bibinfo{year}{2024}.
\newblock \bibinfo{title}{An efficient multi-label classification-based
  municipal waste image identification}.
\newblock \bibinfo{journal}{Processes} \bibinfo{volume}{12},
  \bibinfo{pages}{1075}.
\bibitem[{Xu et~al.(2024)Xu, Tang, Li, Qin and Zou}]{10632565}
\bibinfo{author}{Xu, H.}, \bibinfo{author}{Tang, W.}, \bibinfo{author}{Li, Z.},
  \bibinfo{author}{Qin, K.}, \bibinfo{author}{Zou, J.}, \bibinfo{year}{2024}.
\newblock \bibinfo{title}{Multimodal dual cross-attention fusion strategy for
  autonomous garbage classification system}.
\newblock \bibinfo{journal}{IEEE Transactions on Industrial Informatics}
  \bibinfo{volume}{20}, \bibinfo{pages}{13319--13329}.
\bibitem[{Xu et~al.(2025)Xu, Wang, Zhang and Zhang}]{XU2025107311}
\bibinfo{author}{Xu, Y.}, \bibinfo{author}{Wang, D.}, \bibinfo{author}{Zhang,
  L.}, \bibinfo{author}{Zhang, L.}, \bibinfo{year}{2025}.
\newblock \bibinfo{title}{Dual selective fusion transformer network for
  hyperspectral image classification}.
\newblock \bibinfo{journal}{Neural Networks} \bibinfo{volume}{187},
  \bibinfo{pages}{107311}.
\bibitem[{Yang et~al.(2021)Yang, Zeng, Wang, Zou and Xie}]{9435085}
\bibinfo{author}{Yang, J.}, \bibinfo{author}{Zeng, Z.}, \bibinfo{author}{Wang,
  K.}, \bibinfo{author}{Zou, H.}, \bibinfo{author}{Xie, L.},
  \bibinfo{year}{2021}.
\newblock \bibinfo{title}{Garbagenet: A unified learning framework for robust
  garbage classification}.
\newblock \bibinfo{journal}{IEEE Transactions on Artificial Intelligence}
  \bibinfo{volume}{2}, \bibinfo{pages}{372--380}.
\bibitem[{Zhang et~al.(2024)Zhang, Wang, Gao, Yan and Wang}]{n2}
\bibinfo{author}{Zhang, F.}, \bibinfo{author}{Wang, B.}, \bibinfo{author}{Gao,
  D.}, \bibinfo{author}{Yan, C.}, \bibinfo{author}{Wang, Z.},
  \bibinfo{year}{2024}.
\newblock \bibinfo{title}{When grey model meets deep learning: A new hazard
  classification model}.
\newblock \bibinfo{journal}{Information Sciences} \bibinfo{volume}{670},
  \bibinfo{pages}{120653}.
\bibitem[{Zhang et~al.(2021a)Zhang, Yang, Zhang, Bao, Su and Liu}]{8}
\bibinfo{author}{Zhang, Q.}, \bibinfo{author}{Yang, Q.},
  \bibinfo{author}{Zhang, X.}, \bibinfo{author}{Bao, Q.}, \bibinfo{author}{Su,
  J.}, \bibinfo{author}{Liu, X.}, \bibinfo{year}{2021}a.
\newblock \bibinfo{title}{Waste image classification based on transfer learning
  and convolutional neural network}.
\newblock \bibinfo{journal}{Waste Management} \bibinfo{volume}{135},
  \bibinfo{pages}{150--157}.
\bibitem[{Zhang et~al.(2021b)Zhang, Chen, Yang and Gong}]{5}
\bibinfo{author}{Zhang, S.}, \bibinfo{author}{Chen, Y.}, \bibinfo{author}{Yang,
  Z.}, \bibinfo{author}{Gong, H.}, \bibinfo{year}{2021}b.
\newblock \bibinfo{title}{Computer vision based two-stage waste
  recognition-retrieval algorithm for waste classification}.
\newblock \bibinfo{journal}{Resources, Conservation and Recycling}
  \bibinfo{volume}{169}, \bibinfo{pages}{105543}.
\bibitem[{Zoph et~al.(2018)Zoph, Vasudevan, Shlens and Le}]{zoph2018learning}
\bibinfo{author}{Zoph, B.}, \bibinfo{author}{Vasudevan, V.},
  \bibinfo{author}{Shlens, J.}, \bibinfo{author}{Le, Q.V.},
  \bibinfo{year}{2018}.
\newblock \bibinfo{title}{Learning transferable architectures for scalable
  image recognition}, in: \bibinfo{booktitle}{Proceedings of the IEEE
  conference on computer vision and pattern recognition}, pp.
  \bibinfo{pages}{8697--8710}.

\end{thebibliography}

\bibliographystyle{elsarticle-num}

\end{document}